\documentclass[11pt]{article}

\usepackage[preprint]{acl}

\usepackage{times}
\usepackage{latexsym}
\usepackage[T1]{fontenc}
\usepackage[utf8]{inputenc}
\usepackage{microtype}
\usepackage{inconsolata}
\usepackage{subcaption}

\usepackage{graphicx}
\usepackage{subcaption}
\usepackage{multirow}
\usepackage{booktabs}
\usepackage{placeins}

\usepackage{amsmath}
\usepackage{amssymb}
\usepackage{mathtools}
\usepackage{amsthm}
\usepackage{bm}

\usepackage{xcolor}
\usepackage{soul}
\usepackage{tcolorbox}
\usepackage{algorithm}
\usepackage{algpseudocode}

\definecolor{PVFgreen}{HTML}{008B72}
\definecolor{PVFred}{HTML}{C00000}

\newcommand{\accpos}[1]{\textcolor{PVFgreen}{(+#1)}}
\newcommand{\accneg}[1]{\textcolor{PVFred}{(-#1)}}
\newcommand{\speedup}[1]{\textcolor{PVFgreen}{(#1$\times$)}}

\definecolor{PVFgreen}{HTML}{008B72}
\definecolor{PVFred}{HTML}{C00000}

\newcommand{\metriccell}[2]{%
\begin{tabular}{@{}c@{}}
#1\\[-1pt]
{\footnotesize #2}
\end{tabular}
}

\usepackage[capitalize,noabbrev]{cleveref}

\usepackage[textsize=tiny]{todonotes}

\theoremstyle{plain}

\theoremstyle{definition}

\theoremstyle{remark}

\definecolor{priorityColor}{HTML}{FFD700} 
\definecolor{fallbackColor}{HTML}{FF9999} 

\newcommand{\tokenP}[1]{\colorbox{priorityColor}{\strut #1}\allowbreak}
\newcommand{\tokenF}[1]{\colorbox{fallbackColor}{\strut #1}\allowbreak}
\newcommand{\tokenH}[1]{\colorbox{white}{\strut #1}\allowbreak}

\title{Plan, Verify and Fill: A Structured Parallel Decoding Approach for Diffusion Language Models}

\author{
  \textbf{Miao Li}\textsuperscript{1,*}
  \quad
  \textbf{Hanyang Jiang}\textsuperscript{1,*}
  \quad
  \textbf{Sikai Cheng}\textsuperscript{1}
  \quad
  \textbf{Hengyu Fu}\textsuperscript{2}
  \quad
  \textbf{Yuhang Cai}\textsuperscript{2}
  \\
  \textbf{Baihe Huang}\textsuperscript{2}
  \quad
  \textbf{Zhongzhu Chen}\textsuperscript{3}
  \quad
  \textbf{Tinghan Ye}\textsuperscript{1}
  \quad
  \textbf{Xuanzhou Chen}\textsuperscript{1}
  \quad
  \textbf{Pascal Van Hentenryck}\textsuperscript{1}
  \\
  \textsuperscript{1}Georgia Institute of Technology
  \quad
  \textsuperscript{2}University of California, Berkeley
  \quad
  \textsuperscript{3}University of Michigan
  \quad
  \\
  \textsuperscript{*}Equal contribution
  \\
  \texttt{\{mli746, scottjhy\}@gatech.edu}
}

\begin{document}
\maketitle
\begin{abstract}
Diffusion Language Models (DLMs) enable parallel text generation by denoising multiple masked positions in a single pass, but confidence-threshold decoding often commits only the safest predictions, leaving informative uncertainty underused. We propose \textbf{Plan-Verify-Fill (PVF)}, a training-free draft-and-verification decoding strategy that preserves the high-confidence decoding backbone while selectively testing lower-confidence candidates with high expected verified utility. PVF uses a calibrated planning-token vocabulary to allocate its limited verification budget, accepts proposed drafts only when they remain consistent with the backbone trajectory, and otherwise defaults to conservative confidence-gated unmasking. Across benchmarks spanning mathematical reasoning, knowledge, and code generation, PVF preserves final accuracy while reducing decoding cost. Measured on LLaDA-8B-Instruct, PVF achieves up to $2.3\times$ higher throughput than confidence-threshold decoding.
\end{abstract}
\section{Introduction}
DLMs \citep{gong2024scaling, shi2024simplified,lou2024discrete, ye2025dream, nie2025large} extend the iterative denoising paradigm of diffusion models to discrete text generation, offering an alternative to the strictly sequential decoding employed by AR models. By operating over a global canvas and iteratively refining token predictions, DLMs unlock substantial potential for parallel token updates. 
Recent studies, together with real-world deployments \citep{khanna2025mercury,nie2025large, deepmind2025geminidiffusion}, demonstrate that masked diffusion architectures scale well and deliver competitive performance across a range of tasks. 

The appeal of DLMs lies in their ability to generate in parallel \citep{prabhudesai2025diffusion, chen2024diffusion,arriola2025block}: a single denoising pass predicts many masked positions, enabling multiple tokens to be committed per iteration rather than decoding strictly left-to-right as in autoregressive models. To realize this parallelism while maintaining generation accuracy, existing strategies commonly commit only sufficiently high-confidence tokens \citep{wu2025fast, nie2025large}. This conservative gate is reliable, but it can limit the achievable tokens-per-forward-pass (TPF). We hypothesize that this limitation partly reflects trajectory multiplicity: before surrounding context is resolved, a masked position may remain compatible with several coherent continuations, causing the model to distribute probability mass across alternatives. Consequently, confidence-threshold decoding may underutilize the parallel capacity of each denoising pass, making inference-time decoding a key bottleneck for unlocking the speed potential of DLMs.

A complementary line of work instead exploits uncertainty for speed, relaxing confidence-greedy decoding to expose informative high-entropy or low-confidence positions earlier \citep{ben2025accelerated, kim2026klass, fu2025bits}. While this can reduce decoding rounds, it also sharpens the accuracy--efficiency trade-off: the positions that accelerate convergence are often less reliable commitments, and premature unmasking can propagate errors or constrain later denoising steps. This raises the central question of our work:
\begin{center}
\emph{Can fast DLM decoding move beyond confidence-greedy commitment and instead perform selective exploration that is both informative and reliable?}
\end{center}
More concretely, in addition to committing high-confidence tokens, can we identify lower-confidence candidates whose early unmasking provides useful planning information while remaining compatible with the model's current high-confidence denoising trajectory?

Motivated by this question, we observe that the risk of early unmasking is not uniform across tokens. Confidence-threshold decoding already provides a reliable backbone: high-confidence tokens can be safely committed. The opportunity lies in the lower-confidence region, where some tokens are too uncertain for ordinary confidence-based decoding but still useful for accelerating the denoising trajectory. To exploit this asymmetry, we propose \textbf{Plan-Verify-Fill (PVF)}, a training-free draft-and-verification strategy that preserves the high-confidence backbone while selectively testing a few lower-confidence drafts. PVF commits a draft only when verification shows it remains compatible with the model's current denoising trajectory; otherwise, it falls back to safer local fills. Since verification is useful only within a small efficiency-preserving budget, PVF prioritizes candidates that offer both downstream benefit and a high chance of passing verification.

PVF operationalizes this principle through two complementary routes. The \textbf{Planning Route} uses this separation to look beyond the high-confidence backbone and identify which lower-confidence tokens are worth verifying. Rather than exploring the lower-confidence pool uniformly, it prioritizes candidates with high expected verified utility: tokens whose early unmasking can reduce downstream uncertainty while remaining compatible with the model's current high-confidence base trajectory. Such candidates can resolve ambiguities that confidence-based decoding would otherwise defer, but their lower confidence makes direct commitment risky. PVF therefore treats them as drafts, accepting them only after impact-set checks confirm that the proposed commitment is supported by the predictions it is expected to influence. This targeted proposal process turns limited verification attempts into effective decoding progress.

When reliable planning candidates are unavailable, the \textbf{AR Fallback Route} switches to a safer local strategy: it proposes locally coherent fills that densify the context for later rounds, but accepts them only when they pass an immediate-context consistency check. Thus, PVF reframes fast DLM decoding as verified selective exploration, choosing between global planning and local filling according to where each route offers the most favorable risk--benefit trade-off.

Empirically, we evaluate PVF on four benchmark datasets using LLaDA-8B-Instruct and Dream-7B-Instruct, demonstrating robustness across two representative DLM backbones. Our contributions can be summarized as follows:
\begin{itemize}
\item We introduce \textbf{PVF}, a novel training-free parallel decoding strategy that treats DLM acceleration as verification-aware selective exploration rather than reactive high-confidence token harvesting.
\vspace{-0.1cm}
\item We design a \textbf{dual-route verified commitment} mechanism that spends the limited verification budget selectively: a Planning Route pursues high-value and verifiable drafts, while an AR Fallback Route commits safer local fills when no such planning opportunity is available. 
\vspace{-0.1cm}
\item We demonstrate that PVF consistently improves decoding efficiency while preserving accuracy, showing gains over strong baselines across multiple DLMs and benchmarks.
\end{itemize}

\subsection{Related Work}
\paragraph{Draft-and-Verification-Based Decoding.}
Speculative decoding is a widely adopted family of efficient LLM inference methods built around proposal--verification: a drafter proposes candidate continuations, and a target model verifies them in parallel; self-speculative variants use the same model for both roles \citep{leviathan2023fast,chen2023accelerating,zhang2024draft}. Its gains depend on a favorable acceptance regime, where accepted candidates amortize the verifier call. For DLM-style batched verification, recent work further identifies a narrow ``free-lunch'' region: a small number of candidates can be checked with little additional wall-clock latency because small-batch GPU inference can remain memory-bandwidth limited, but larger verification width increases latency, memory pressure, and wasted computation from rejected candidates \citep{wu2025free,fu2025bits}. 

\paragraph{Verification-Guided DLM Decoding.} Recent DLM methods adapt this principle to masked denoising: FreeDAVE verifies candidates along the static DLM trajectory, SSD self-drafts masked positions with hierarchical verification, and WINO drafts broadly but re-masks suspicious tokens under stricter verification \citep{wu2025free,gao2025self,hong2025wide}. These methods enlarge the safe acceptance window, but primarily organize verification around agreement with the high-confidence base trajectory. PVF instead treats the limited free-lunch budget as verification-aware selective exploration, spending it on candidates with high expected verified utility: informative planning tokens when they are likely to pass impact-set checks, and safer AR-style fallback fills otherwise.

\paragraph{Token-level Planning.}
The limited low-overhead verification regime in proposal--verification decoding puts pressure on candidate selection for DLMs: unlike AR decoding, which extends a single left-to-right prefix, masked decoding exposes a combinatorial space of positions and generation orders, so useful drafts should be both likely to pass and worth revealing. Recent analyses of reasoning traces suggest that the downstream utility of revealing a position is highly non-uniform: some high-entropy positions are not merely uncertain residues, but trajectory-defining choices that can shape the remaining generation \citep{wang2025beyond,cheng2025reasoning,ni2026flexibility}. 

Purely confidence-based DLM decoding can therefore be too locally conservative: by filling easy tokens first, it may bypass high-entropy decision points, and \citet{ni2026flexibility} shows that such bypassing hurts pass@$k$ accuracy compared with decoding orders that confront these positions directly. PVF adapts this insight to inference-time decoding by treating selected high-entropy positions as verifiable drafts: instead of globally lowering the confidence threshold, which can trade accuracy for TPF \citep{wu2025fast}, PVF targets uncertain candidates with high expected information gain and accepts them only when they remain compatible with the current denoising consensus, turning limited verification budget into reliable progress before their ambiguity is eliminated by surrounding context.
\section{Preliminary}
\label{sec:Preliminary}
\begin{figure*}[t]
    \centering

    \begin{subfigure}[t]{0.45\linewidth}
        \centering
        \includegraphics[width=\linewidth]{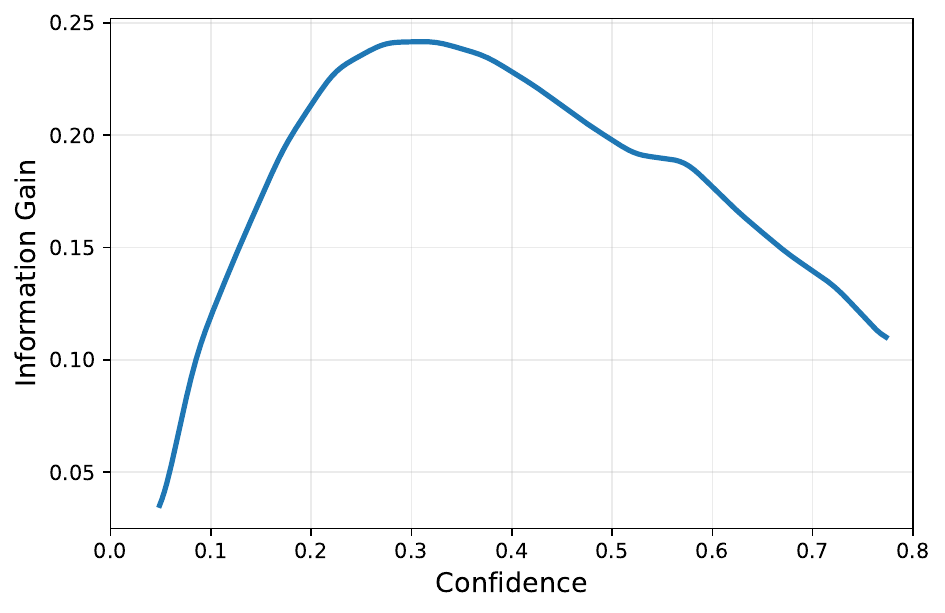}
        \caption{Average over all candidate tokens.}
        \label{fig:confidence-information-gain-all}
    \end{subfigure}
    \hfill
    \begin{subfigure}[t]{0.45\linewidth}
        \centering
        \includegraphics[width=\linewidth]{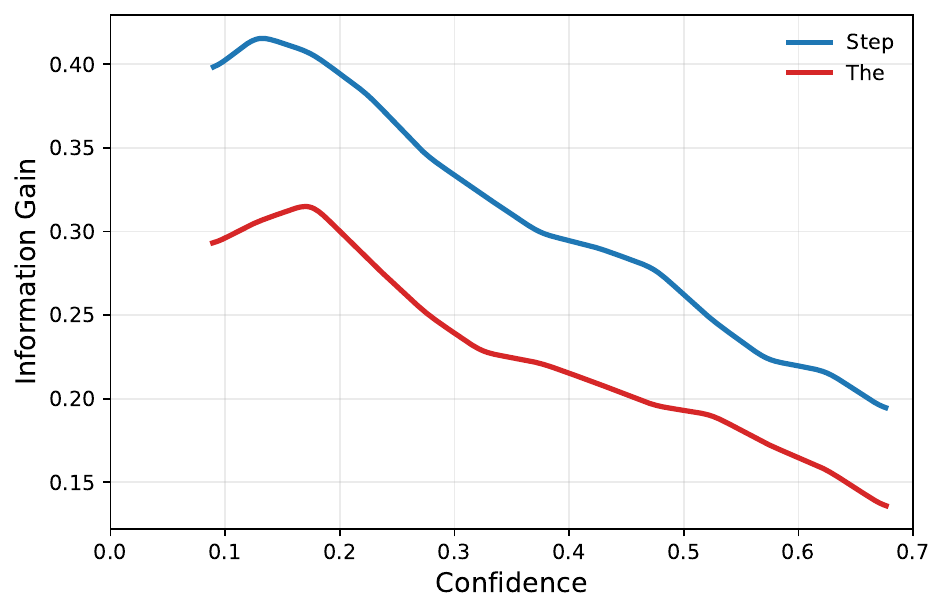}
        \caption{Representative tokens \texttt{Step} and \texttt{The}.}
        \label{fig:confidence-information-gain-token}
    \end{subfigure}

    \vspace{-0.2cm}
    \caption{
    Relationship of information gain of candidate token commitments versus model confidence on GSM8K.
    \textbf{(a)} Average over all candidate tokens, showing a sweet spot in the low to moderate confidence region where candidates provide high marginal information gain.
    \textbf{(b)} Two representative vocabulary tokens, \texttt{Step} and \texttt{The}, showing that information gain varies by token even at similar confidence.
    }
    \label{fig:confidence-information-gain}
    \vspace{-0.3cm}
\end{figure*}
Departing from the left-to-right decoding of AR models, DLMs
\cite{austin2021structured, lou2023discrete, ou2024your} formulate generation as iterative denoising. They define a forward corruption process that gradually masks a clean sequence, and learn a reverse process that reconstructs text from partially masked states.

\paragraph{Forward Process.}
Given a clean sequence $\mathbf{x}_0$, the forward process independently corrupts each token over continuous time $s\in[0,1]$ by replacing it with a special absorbing token \texttt{[MASK]}. Once masked, a token remains masked, so $\mathbf{x}_1$ is fully masked. Following \cite{austin2021structured}, this process is
{\setlength{\abovedisplayskip}{0pt}
\setlength{\belowdisplayskip}{0pt}
\begin{align*}
q(\mathbf{x}_s|\mathbf{x}_0)
&= \prod_{i=1}^{|\mathbf{x}_0|} q(x_s^i|x_0^i) \\
&= \prod_{i=1}^{|\mathbf{x}_0|}
\mathrm{Cat}\!\left(x_s^i;(1-s)\delta_{x_0^i}+s\delta_{\texttt{[MASK]}}\right).
\end{align*}}
\paragraph{Reverse Process.} Given $\mathbf{y}=[\mathbf{y}_{\mathrm{prompt}},\mathbf{y}_{\mathrm{gen}}]$, where $\mathbf{y}_{\mathrm{prompt}}$ is fixed and $\mathbf{y}_{\mathrm{gen}}$ is initialized as masked tokens. A denoising model $P_\theta$ is trained typically with an ELBO-based objective, to recover clean tokens from corrupted states \cite{nie2025large, sahoo2024simple}. At inference time, generation iteratively applies $P_\theta(\cdot|\mathbf{y}_t)$ to a partially masked sequence $\mathbf{y}_t$, producing token distributions over the vocabulary for all masked positions in parallel. A decoding rule then selects which predictions to commit, yielding a progressively less-masked sequence until generation completes.

\paragraph{Confidence-based Parallel Decoding:}
One of the most prominent decoding strategies in DLMs is confidence-based decoding~\cite{yu2025dimple}, which governs the generation process based on prediction certainty. For each masked position $i$ at step $t$, define the model's top prediction $\hat{y}_{t}^i$ conditioned on the previous state $\mathbf{y}_{t-1}$:
\[
\hat{y}_{t}^i = \operatorname*{arg\,max}_{w \in \mathcal{V}} P_\theta(y^i = w \mid \mathbf{y}_{t-1}).
\]
{In its most canonical form}, the algorithm progressively fills the sequence by committing to tokens where the model's confidence exceeds a static threshold $\tau_{\text{high}}$: 
\begin{align*}\label{eqn: threshold-based parallel}
    y_t^i = \begin{cases}
        \hat{y}_{t}^i, & \text{if } P_\theta(y^i = \hat{y}_{t}^i \mid \mathbf{y}_{t-1}) \geq \tau_{\text{high}}, \\
        y_{t-1}^i, & \text{otherwise}.
    \end{cases}
\end{align*}

While this static heuristic serves as a basic building block for parallel decoding, later approaches~\cite{chen2025dparallel,jin2025thinking,lou2024discrete} adopt more complex decoding mechanisms that extend beyond heuristic gating.

\section{Selecting High-Entropy Candidates for Planning}
\label{sec:planning-token-intuition}

Confidence-based parallel decoding provides a reliable base trajectory by committing positions whose confidence exceeds $\tau_{\mathrm{high}}$. Our PVF preserves this base behavior: high-confidence tokens are committed by default and serve as the backbone of the current step. However, because these tokens are already near-certain, their marginal information value is limited. As Figure~\ref{fig:confidence-information-gain-all} shows, candidate commitments in a low to moderate confidence region can yield larger information gain than high-confidence tokens, while extremely low-confidence candidates are less useful. This suggests that the main acceleration opportunity lies in carefully selected uncertain positions that help resolve the surrounding trajectory earlier.

Figure~\ref{fig:confidence-information-gain-token} further shows that information gain is not determined by confidence alone. On GSM8K, \texttt{Step} frequently appears in step-by-step solutions such as \texttt{Step 1} and \texttt{Step 2}, and yields consistently higher information gain than a generic token such as \texttt{The}. This motivates selecting a compact proposal set of uncertain candidates that are both compatible with the high-confidence trajectory and informative for the remaining masked positions. Since only a few candidates can be checked within the small-batch free-lunch regime, we use a planning-token vocabulary $P^{\mathrm{plan}}$ to restrict the proposal space. This vocabulary only controls eligibility. The full PVF algorithm still verifies each proposal online before accepting it.

\paragraph{Base branch and impact set.}
We use the high-confidence decoding path as a local reference for evaluating riskier high-entropy commitments. Intuitively, the base set contains the positions that confidence-only decoding would safely commit at the current step. The base branch is the resulting sequence after making only these safe commits. It is not ground truth, but it is the most reliable local trajectory available to a training-free decoder. A planning draft should therefore preserve the future predictions that this base path already makes with high confidence.

Let $B_t$ denote the active block at step $t$, and let
\[
M_t
=
\{i\in B_t \mid y_{t-1}^i=\texttt{[MASK]}\}
\]
be the currently masked positions. For each $i\in M_t$, define the top prediction and its confidence as
\[
\begin{aligned}
\hat y_t^i
&=
\arg\max_{w\in V}
P_\theta(y^i=w\mid \mathbf{y}_{t-1}),
\\
p_t^i
&=
P_\theta(y^i=\hat y_t^i\mid \mathbf{y}_{t-1}).
\end{aligned}
\]
The base set and base branch are
\[
S_t^{\mathrm{base}}=
\{i\in M_t \mid p_t^i\ge \tau_{\mathrm{high}}\},
\]
\[
z_t^{i,\mathrm{base}}
=
\begin{cases}
\hat y_t^i, & i\in S_t^{\mathrm{base}},\\
y_{t-1}^i, & i\notin S_t^{\mathrm{base}}.
\end{cases}
\]
We define the impact set as the positions that remain masked after the base commit but become high-confidence under the base branch:
\begin{equation}
S_t^{\mathrm{impact}}
=
\left\{
i\in B_t
\,\middle|\,
\begin{aligned}
&z_t^{i,\mathrm{base}}=\texttt{[MASK]},\\
&\max_{w\in V}
P_\theta(y^i=w\mid \mathbf{z}_t^{\mathrm{base}})
\ge \tau_{\mathrm{high}}
\end{aligned}
\right\}.
\label{eq:impact_set}
\end{equation}

These positions form the high-confidence future that additional commitments should preserve.

\paragraph{Verified compatibility.}
A high-entropy candidate should not disturb the high-confidence path established by the base branch. For candidate $j$, let $i_{(j)}$ denote its position. We form a temporary planning branch by starting from the base branch and additionally committing this candidate:
\begin{equation}
z_t^{i,\mathrm{plan},j}
=
\begin{cases}
\hat y_t^{i_{(j)}}, & i=i_{(j)},\\
z_t^{i,\mathrm{base}}, & \text{otherwise}.
\end{cases}
\label{eq:plan_branch}
\end{equation}

This planning branch is only a draft. It is not committed unless it passes verification.

The candidate is verified-compatible if the impact set is nonempty and the high-confidence top predictions on all positions in $S_t^{\mathrm{impact}}$ remain the same under $\mathbf{z}_t^{\mathrm{plan},j}$ as under $\mathbf{z}_t^{\mathrm{base}}$. We denote this binary outcome by $A_t^j\in\{0,1\}$:
\begin{equation}
A_t^j=1
\quad\Longleftrightarrow\quad
\begin{aligned}[t]
&|S_t^{\mathrm{impact}}|>0
\text{ and}\\
&\mathbf{z}_t^{\mathrm{plan},j}
\text{ preserves top-1}\\
&\text{base predictions on }S_t^{\mathrm{impact}} .
\end{aligned}
\label{eq:at}
\end{equation}

Thus, $A_t^j=1$ means that the candidate is locally consistent with the model's high-confidence reference path. This is not a guarantee of global correctness. It is a lightweight stability check used to avoid commitments that immediately contradict the base trajectory.

\paragraph{Verified information gain.}
Verified compatibility screens for risk, but it does not measure usefulness. Since high-confidence tokens are already committed by the base decoder, useful additional commitments should make the unresolved part of the block less uncertain. We therefore score a candidate by the entropy reduction it produces over the remaining masked positions after passing verification.

Let
\[
H_i(\mathbf{z})
=
H\!\left(P_\theta(y^i\mid \mathbf{z})\right)
\]
be the entropy of the model's token distribution at position $i$ under state $\mathbf{z}$. For candidate $j$, define the evaluation region as
\[
R_t^j
=
\left\{
i\in B_t
\,\middle|\,
z_t^{i,\mathrm{base}}=\texttt{[MASK]},
\ 
i\neq i_{(j)}
\right\}.
\]
This excludes the candidate position itself, so the score measures how the candidate affects the remaining masked positions. Define
\[
d_{t,i}^j
=
H_i(\mathbf{z}_t^{\mathrm{base}})
-
H_i(\mathbf{z}_t^{\mathrm{plan},j}).
\]
When $R_t^j$ is nonempty, the entropy reduction induced by candidate $j$ is
\[
\Delta H_t^j
=
\frac{1}{|R_t^j|}
\sum_{i\in R_t^j}
d_{t,i}^j.
\]
If $R_t^j$ is empty, we set $\Delta H_t^j=0$. The event-level verified information gain is
\[
\mathrm{VIG}_t^j
=
A_t^j \Delta H_t^j.
\]
A positive $\mathrm{VIG}_t^j$ means that the candidate both preserves the high-confidence path and reduces uncertainty over the remaining block.

\paragraph{Planning-token vocabulary.}
We compute verified compatibility and verified information gain for each vocabulary token on an offline calibration set. Tokens that appear often enough and achieve high average scores on both metrics are retained as the planning-token vocabulary $P^{\mathrm{plan}}$. This selection is data-driven rather than manually specified: it identifies tokens that are empirically likely to pass verification and reduce uncertainty when proposed in high-entropy positions. The resulting vocabulary is frozen before evaluation and used only as an eligibility filter for the planning route. Every proposed commitment is still verified online by PVF. The full calibration procedure, thresholds, and data sources are provided in Appendix~\ref{sec:planning tokens}.

\begin{figure*}[H]
    \centering    \includegraphics[width=0.9\linewidth]{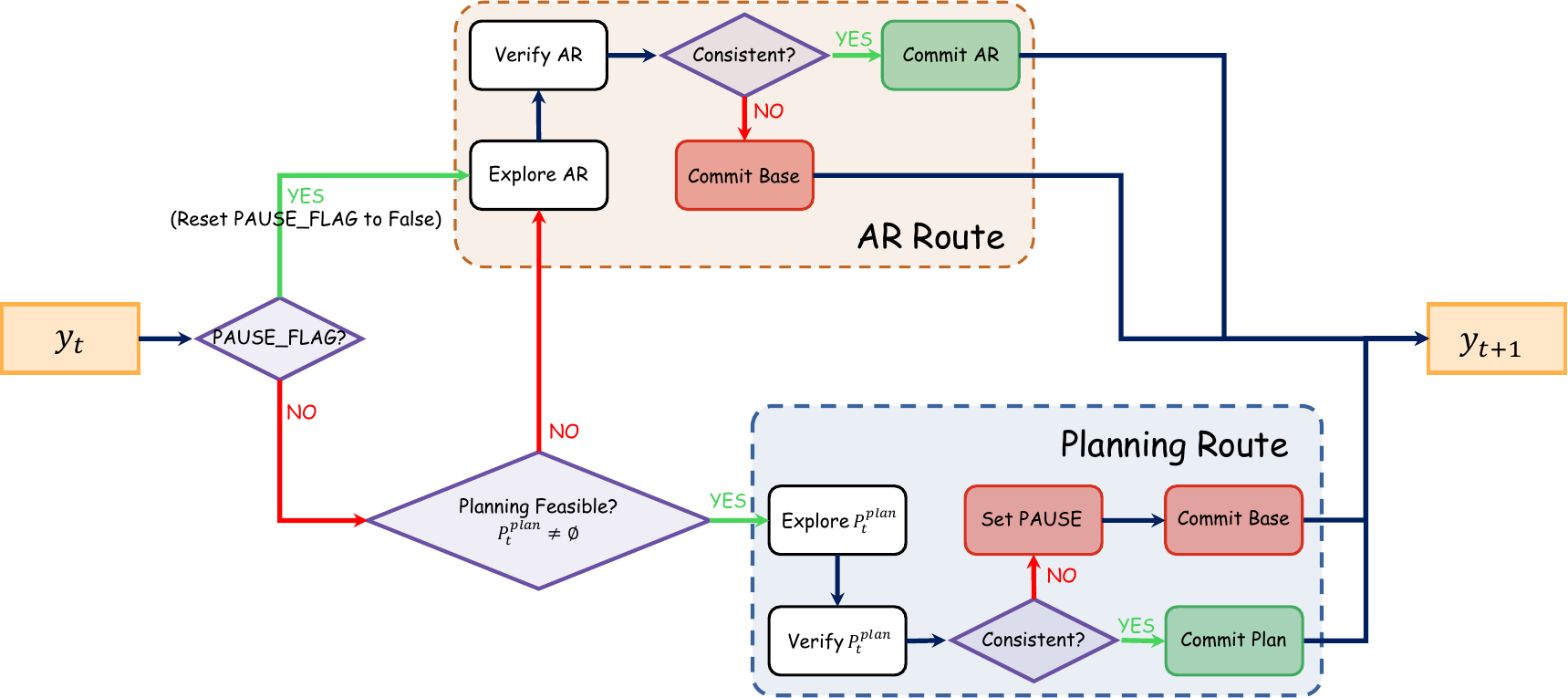}
\caption{Overview of the \textbf{Plan--Verify--Fill (PVF)} decoding pipeline.}
\label{fig:pvf_overview}
\end{figure*}
\begin{figure*}[t]
    \centering    \includegraphics[height=7cm, width=\linewidth ]{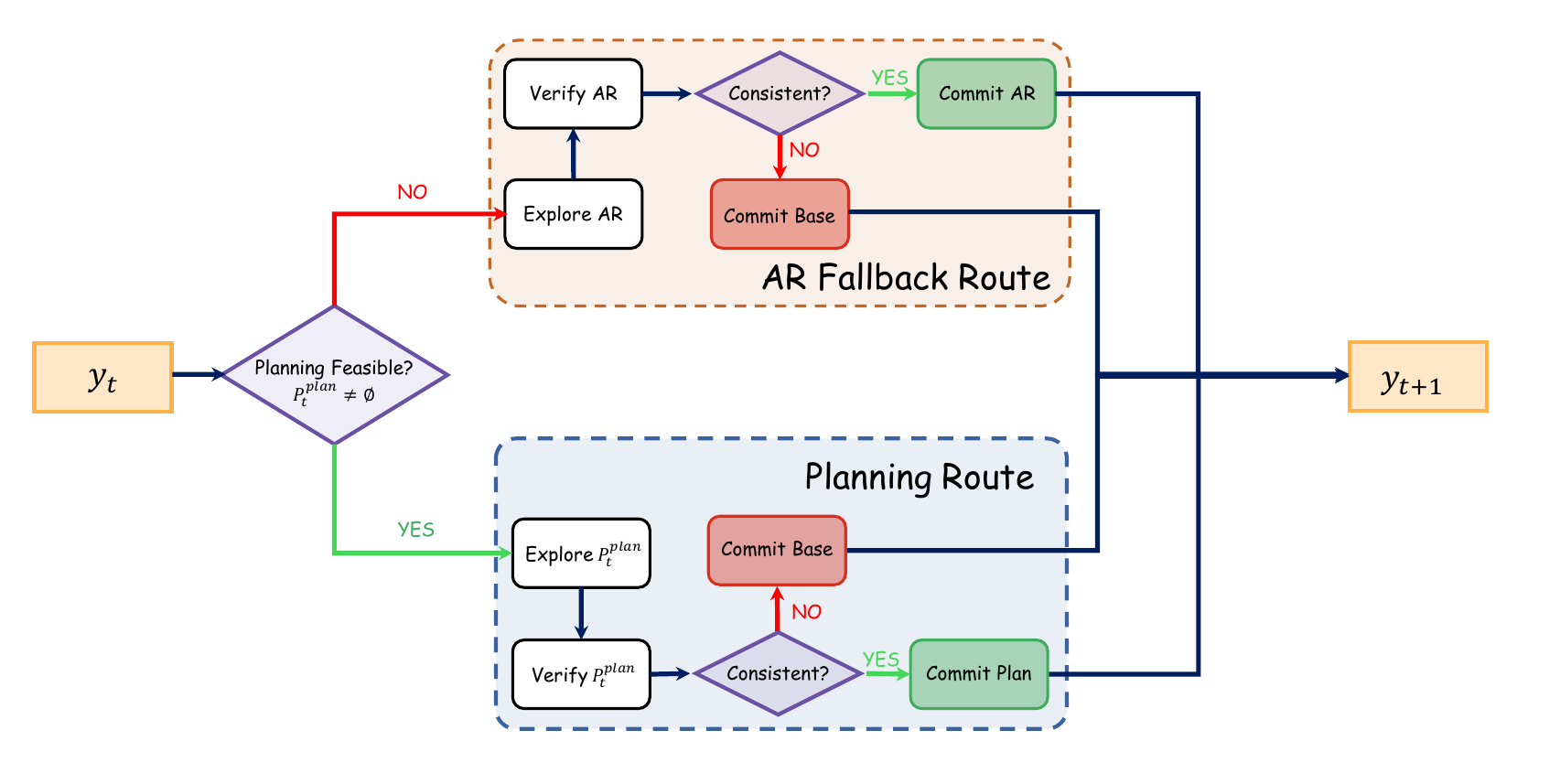}
\caption{Overview of the \textbf{Plan--Verify--Fill (PVF)} decoding pipeline.}
\label{fig:pvf_overview}
\end{figure*}

\section{Methodology: Plan-Verify-Fill (PVF) Decoding}
\label{sec:method}

We propose \textbf{Plan-Verify-Fill (PVF)} (Figure \ref{fig:pvf_overview}), a training-free draft-and-verification decoder for DLMs. PVF keeps confidence-threshold decoding as a high-confidence backbone: at each decoding step $t$, it forms $\mathbf{z}_t^{\mathrm{base}}$ from $S_t^{\mathrm{base}}$, then uses $S_t^{\mathrm{impact}}$ defined in Equation~\eqref{eq:impact_set} to verify a small number of additional drafts. These drafts come from two routes: the \textbf{Planning Route}, which targets lower-confidence candidates from $P^{\mathrm{plan}}$, and the \textbf{AR Fallback Route}, which supplies safer local fills when planning is unavailable.

\subsection{Planning Route Phase I: Candidate Proposal}

PVF exploits the small-batch verification regime by testing only a few additional drafts alongside the base branch. The Planning Route selects these exploratory drafts from lower-confidence candidates whose predicted token lies in $P^{\mathrm{plan}}$, treating them as proposals that are committed only after impact-set verification supports their compatibility with the current denoising trajectory.

\paragraph{Planning candidates.}
Using the planning-token set $P^{\mathrm{plan}}$ constructed in Section~\ref{sec:planning-token-intuition}, PVF selects non-base masked positions whose top prediction is a planning token and whose confidence falls in a planning band:
\begin{equation*}
\begin{aligned}
P_t^{\mathrm{plan}}
=
\Big\{i\in M_t\setminus S_t^{\mathrm{base}}
\,&\Big|\,
\hat y_t^i\in P^{\mathrm{plan}},\\
&\tau_{\mathrm{plan}}^\ell
\le
p_t^i <\tau_{\mathrm{plan}}^u\Big\}
.    
\end{aligned}    
\end{equation*}
The lower bound filters highly unstable predictions, while the upper bound removes near-confident positions that offer limited additional information. The band therefore targets candidates with meaningful uncertainty but enough model support to be verification-worthy.

PVF keeps the top-$K$ candidates by confidence:
\[
\{i_{(1)},\ldots,i_{(K_t)}\}
=
\operatorname{TopK}_{i\in P_t^{\mathrm{plan}}}(p_t^i),
\]
where $K_t=\min(K,|P_t^{\mathrm{plan}}|)$. For each selected candidate $i_{(j)}$, PVF instantiates the temporary planning branch $\mathbf{z}_t^{\mathrm{plan},j}$ defined in Equation~\eqref{eq:plan_branch}, and evaluates all branches in one batched forward pass:
\[
\left\{
\mathbf{z}_t^{\mathrm{base}},
\mathbf{z}_t^{\mathrm{plan},1},
\ldots,
\mathbf{z}_t^{\mathrm{plan},K_t}
\right\}.
\]
We use $K=3$ by default, keeping verification within the small-width regime used by prior draft-and-verification DLM methods \citep{wu2025free,gao2025self,hong2025wide}. If $P_t^{\mathrm{plan}}=\emptyset$, PVF skips the Planning Route and uses the AR Fallback Route.

\subsection{Planning Route Phase II: Verification and Selection}

The Planning Route separates offline usefulness from online acceptability. The offline VIG score is used only to build $P^{\mathrm{plan}}$, identifying token types that are promising to explore. During decoding, however, every concrete proposal is re-verified in context: PVF accepts a planning draft only if its impact set remains compatible with the current high-confidence trajectory. In this sense, $P^{\mathrm{plan}}$ provides the proposal prior, while online verification provides the commitment test.

\paragraph{Filter 1: Impact-set verification.}
For each planning branch $\mathbf{z}_t^{\mathrm{plan},j}$, PVF computes the verification indicator $A_t^j$ defined in Equation~\eqref{eq:at}. Only verified candidates are preserved:
\[
\mathcal{J}_t^{\mathrm{plan}}
=
\{j\in [K_t] \mid A_t^j=1\}.
\]
Equivalently, a planning branch is accepted by Filter 1 only if it preserves the base branch's top-1 predictions on the nonempty impact set $S_t^{\mathrm{impact}}$. This makes the proposed commitment consistent with the model-internal high-confidence base trajectory used as the reference trajectory. If $\mathcal{J}_t^{\mathrm{plan}}=\emptyset$, PVF treats planning as premature for the current state, and commits $
\mathbf{y}_t \leftarrow \mathbf{z}_t^{\mathrm{base}}.$
 
\paragraph{Filter 2: Runtime confidence maximization.}
If at least one planning branch passes Filter 1, PVF selects among the verified branches using a runtime confidence objective:
\[
\mathcal{C}_{\mathrm{total}}(\mathbf{z})
=
\sum_{\substack{i\in B_t\\ z^i=\texttt{[MASK]}}}
\max_{w\in V}
P_\theta(y^i=w\mid \mathbf{z}).
\]
PVF selects
\[
j^*
=
\arg\max_{j\in \mathcal{J}_t^{\mathrm{plan}}}
\mathcal{C}_{\mathrm{total}}
(\mathbf{z}_t^{\mathrm{plan},j})
\]
and commits $\mathbf{y}_t
\leftarrow
\mathbf{z}_t^{\mathrm{plan},j^*}$.

This final selection favors the verified branch that makes the remaining masked positions in the current decoding horizon most ready for future commitment.

\subsection{AR Fallback Route: Local Filling}
\subsubsection{AR Fallback Route}
\label{sec:ar-fallback}

When no planning draft is available, PVF uses an AR fallback route to make conservative local progress. The fallback proposes short left-to-right extensions from the remaining non-base masked positions and accepts only the longest extension that remains stable under verification.

\paragraph{AR candidate construction.}
Let $i_1^{\mathrm{AR}},\ldots,i_{K_t^{\mathrm{AR}}}^{\mathrm{AR}}$ be the leftmost positions in
$M_t\setminus S_t^{\mathrm{base}}$ whose confidence satisfies $p_t^i\ge\tau_{\mathrm{AR}}^\ell$, ordered by position, where $K_t^{\mathrm{AR}}\le K$. For each $k\in[K_t^{\mathrm{AR}}]$, PVF constructs an AR branch by adding the first $k$ proposed fills to the base branch:
\[
z_t^{i,\mathrm{AR},k}
=
\begin{cases}
\hat y_t^i, 
& i\in\{i_1^{\mathrm{AR}},\ldots,i_k^{\mathrm{AR}}\},\\[2pt]
z_t^{i,\mathrm{base}},
& \text{otherwise}.
\end{cases}
\]
\paragraph{AR verification and selection.}
PVF accepts an AR branch only if its proposed fills remain unchanged under the verified trajectory:
\begin{equation*}
\begin{aligned}
\mathcal{K}_t^{\mathrm{AR}}
=\Big\{
&k\in[K_t^{\mathrm{AR}}]
\,\Big|\,
\forall r\le k,\; \\
&\arg\max_{w\in V}
P_\theta(y^{i_r^{\mathrm{AR}}}=w\mid \mathbf{z}_t^{\mathrm{AR},k})=\hat y_t^{i_r^{\mathrm{AR}}}
\Big\}.
\end{aligned}    
\end{equation*}

If $\mathcal{K}_t^{\mathrm{AR}}\neq\emptyset$, PVF commits the longest verified extension,
\vspace{-0.1cm}
\[
k^*=\max \mathcal{K}_t^{\mathrm{AR}},
\qquad
\mathbf{y}_t\leftarrow \mathbf{z}_t^{\mathrm{AR},k^*}.
\]
Otherwise, it falls back to the base branch, $\mathbf{y}_t\leftarrow\mathbf{z}_t^{\mathrm{base}}$. 

\begin{table*}[t]
\centering
\caption{Accuracy and NFE comparison with baseline methods. Brackets report accuracy change from Static and NFE speedup over Static.}
\label{tab:main_results_acc_nfe}
\small
\setlength{\tabcolsep}{2.7pt}
\renewcommand{\arraystretch}{0.92}
\begin{tabular*}{\textwidth}{@{\extracolsep{\fill}} ll*{8}{c}}
\toprule
\multirow{2}{*}{\textbf{Model}} &
\multirow{2}{*}{\textbf{Method}} &
\multicolumn{2}{c}{\textbf{GSM8K}} &
\multicolumn{2}{c}{\textbf{MMLU-Pro}} &
\multicolumn{2}{c}{\textbf{HumanEval}} &
\multicolumn{2}{c}{\textbf{Math}} \\
\cmidrule(lr){3-4}
\cmidrule(lr){5-6}
\cmidrule(lr){7-8}
\cmidrule(lr){9-10}
& & \textbf{Acc.} & \textbf{NFE $\downarrow$}
  & \textbf{Acc.} & \textbf{NFE $\downarrow$}
  & \textbf{Acc.} & \textbf{NFE $\downarrow$}
  & \textbf{Acc.} & \textbf{NFE $\downarrow$} \\
\midrule

\multirow[c]{8}{*}{\shortstack{\textbf{LLaDA-8B}\\ \textbf{Instruct}}}
& \multirow[c]{2}{*}{Static}
& \multirow[c]{2}{*}{79.46} & \multirow[c]{2}{*}{512.00}
& \multirow[c]{2}{*}{36.80} & \multirow[c]{2}{*}{512.00}
& \multirow[c]{2}{*}{46.95} & \multirow[c]{2}{*}{512.00}
& \multirow[c]{2}{*}{41.06} & \multirow[c]{2}{*}{512.00} \\
& & & & & & & & & \\

& \multirow[c]{2}{*}{Fast-dLLM}
& 79.62 & 89.77
& 36.20 & 80.26
& 47.56 & 124.26
& 40.84 & 147.16 \\
& & \accpos{0.16} & \speedup{5.70}
& \accneg{0.60} & \speedup{6.38}
& \accpos{0.61} & \speedup{4.12}
& \accneg{0.22} & \speedup{3.48} \\

& \multirow[c]{2}{*}{FreeDave}
& 79.08 & 135.65
& 36.70 & 136.36
& 45.73 & 118.72
& 40.90 & 142.82 \\
& & \accneg{0.38} & \speedup{3.77}
& \accneg{0.10} & \speedup{3.75}
& \accneg{1.22} & \speedup{4.31}
& \accneg{0.16} & \speedup{3.58} \\

& \multirow[c]{2}{*}{PVF}
& \textbf{79.75} & \textbf{31.34}
& 36.40 & \textbf{29.55}
& 46.66 & \textbf{70.31}
& 40.24 & \textbf{73.56} \\
& & \accpos{0.29} & \speedup{16.34}
& \accneg{0.40} & \speedup{17.33}
& \accneg{0.29} & \speedup{7.28}
& \accneg{0.82} & \speedup{6.96} \\

\midrule

\multirow[c]{8}{*}{\shortstack{\textbf{Dream-7B}\\ \textbf{Instruct}}}
& \multirow[c]{2}{*}{Static}
& \multirow[c]{2}{*}{78.00} & \multirow[c]{2}{*}{512.00}
& \multirow[c]{2}{*}{41.30} & \multirow[c]{2}{*}{512.00}
& \multirow[c]{2}{*}{60.37} & \multirow[c]{2}{*}{512.00}
& \multirow[c]{2}{*}{42.20} & \multirow[c]{2}{*}{512.00} \\
& & & & & & & & & \\

& \multirow[c]{2}{*}{Fast-dLLM}
& 77.71 & 75.83
& 40.18 & 92.16
& 59.14 & 112.78
& 43.46 & 129.77 \\
& & \accneg{0.29} & \speedup{6.75}
& \accneg{1.12} & \speedup{5.56}
& \accneg{1.23} & \speedup{4.54}
& \accpos{1.26} & \speedup{3.95} \\

& \multirow[c]{2}{*}{FreeDave}
& 79.92 & 141.01
& 41.10 & 91.74
& 59.15 & 112.82
& 42.00 & 130.25 \\
& & \accpos{1.92} & \speedup{3.63}
& \accneg{0.20} & \speedup{5.58}
& \accneg{1.22} & \speedup{4.54}
& \accneg{0.20} & \speedup{3.93} \\

& \multirow[c]{2}{*}{PVF}
& 79.20 & \textbf{32.34}
& \textbf{43.30} & \textbf{35.22}
& \textbf{62.66} & \textbf{50.89}
& \textbf{43.66} & \textbf{64.80} \\
& & \accpos{1.20} & \speedup{15.83}
& \accpos{2.00} & \speedup{14.54}
& \accpos{2.29} & \speedup{10.06}
& \accpos{1.46} & \speedup{7.90} \\

\bottomrule
\end{tabular*}
\end{table*}

\begin{table}[t]
\centering
\caption{TPS comparison on LLaDA-8B-Instruct. Brackets report speedup over Static.}
\label{tab:llada_tps}
\small
\setlength{\tabcolsep}{3.5pt}
\renewcommand{\arraystretch}{1.15}
\begin{tabular}{lcccc}
\toprule
\textbf{Dataset} & \textbf{Static} & \textbf{Fast-dLLM} & \textbf{FreeDave} & \textbf{PVF} \\
\midrule
GSM8K
& 31.74
& \metriccell{111.37}{\speedup{3.51}}
& \metriccell{71.24}{\speedup{2.24}}
& \metriccell{\textbf{259.35}}{\speedup{8.17}} \\
MMLU-Pro
& 30.48
& \metriccell{64.20}{\speedup{2.11}}
& \metriccell{53.89}{\speedup{1.77}}
& \metriccell{\textbf{113.78}}{\speedup{3.73}} \\
HumanEval
& 30.25
& \metriccell{100.04}{\speedup{3.31}}
& \metriccell{108.10}{\speedup{3.57}}
& \metriccell{\textbf{163.26}}{\speedup{5.40}} \\
Math
& 32.41
& \metriccell{96.75}{\speedup{2.99}}
& \metriccell{87.34}{\speedup{2.69}}
& \metriccell{\textbf{161.12}}{\speedup{4.97}} \\
\bottomrule
\end{tabular}
\end{table}

\section{Experiment Results}

\begin{figure}
    \centering
    \includegraphics[width=\linewidth]{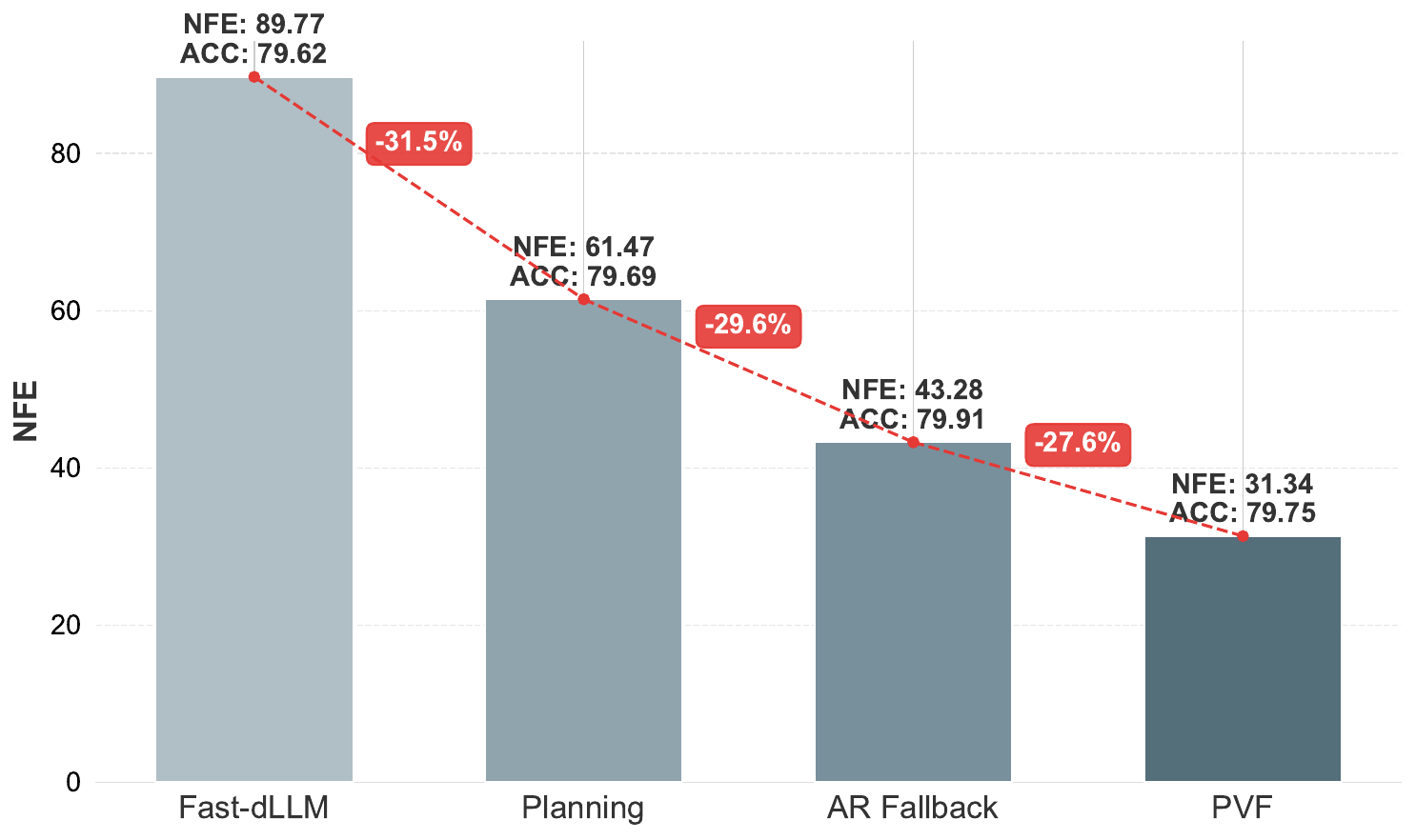}
    \caption{Ablation study on GSM8k, evaluating the contribution of each PVF component. Accuracy scores are displayed in the bracket above each bar to confirm they remain comparable across methods.}
    \label{fig:ablation_component}
    
\end{figure}

\subsection{Experiment Settings}
We evaluate PVF on four benchmarks datasets spanning mathematical reasoning, coding synthesis, and broad knowledge: GSM8K \cite{cobbe2021training}, MMLU-Pro \cite{wang2024mmlupro}, HumanEval \cite{chen2021evaluating}, and MATH \cite{hendrycks2021measuring}. LLaDA-8B-Instruct and Dream-7B-Instruct are used as base models to verify that PVF’s performance gains are robust across two representative backbones. We report both accuracy and efficiency, using the Number of Function Evaluations (NFE), the total number of model forward passes required to complete a sequence as our main efficiency metric. Besides, we also report tokens generated per second (TPS) under KV cache \cite{wu2025fast} as the realized throughput metric, allowing us to verify whether fewer forward passes translate into faster generation. To ensure a fair comparison, we fix $\tau_{\text{high}}=0.9$ across all experiments unless stated otherwise. Unless otherwise specified, we follow \cite{fu2025bits,wu2025free} and use a batch size of $4$ ($k=3$), which lies within the ``free lunch'' region.

Across all datasets, we use a fixed maximum generation length of $L=512$ and a block size of $B=64$. 
Decoding is training-free (no finetuning, no additional supervision, and no external verifier). Sensitivity to $L$ and $B$ is reported in Section~\ref{sec:Ablation Study: Sensitivity to Batch Size}.
All experiments are conducted on a single NVIDIA H200 GPU. Additional experimental details are deferred to section \ref{sec:Additional Experimental Details}.

\subsection{Main Results}\label{subsec: Main Results}
\paragraph{Baselines.}
We compare PVF with three training-free DLM decoding baselines. \textbf{Static} greedily commits the most confident token at each iteration. \textbf{Fast-dLLM}~\cite{wu2025fast} performs confidence-threshold parallel decoding by committing all tokens whose posterior confidence exceeds $\tau_{\mathrm{high}}$. \textbf{FreeDave}~\cite{wu2025free} follows a draft-and-verify design, generating look-ahead drafts and committing the longest consecutively verified span under top-1 decoding to reduce the number of model forward passes. Additional experiment results are deferred to section~\ref{sec:add_exp}.

\paragraph{Accuracy and NFE.}
Table~\ref{tab:main_results_acc_nfe} summarizes accuracy and NFE across four benchmarks. PVF consistently reduces forward passes while preserving comparable accuracy. Relative to Fast-dLLM, PVF achieves a $1.8\times$--$2.9\times$ NFE speedup across datasets and model backbones, corresponding to roughly $43\%$--$65\%$ fewer forward passes. The accuracy changes remain small: on LLaDA-8B-Instruct, PVF is within one point of Fast-dLLM on all benchmarks, and on Dream-7B-Instruct, PVF improves accuracy on all four benchmarks. Compared with FreeDave, PVF also consistently uses fewer forward passes, showing that selective planning and AR fallback provide additional savings beyond draft-and-verify decoding alone.

\paragraph{Throughput.}
NFE is an algorithmic efficiency metric, but a lower NFE is useful only if it translates into actual throughput. Table~\ref{tab:llada_tps} confirms that PVF improves TPS over Fast-dLLM on every benchmark, corresponding to roughly $1.6$--$2.3\times$ higher throughput than Fast-dLLM. This indicates that the additional verification work does not erase the gains from fewer decoding rounds. Within the small-batch regime, the NFE reduction is reflected in real generation speed.

\paragraph{Ablation study.}
Figure~\ref{fig:ablation_component} decouples the two PVF routes. Planning-only is activated only when eligible planning candidates exist, whereas AR Fallback can operate at every decoding round, giving it more frequent opportunities to add verified tokens and explaining its lower standalone NFE. Still, Planning reduces NFE from $89.77$ to $61.47$ with comparable accuracy, showing that high-quality planning in the low-confidence region is valuable when available. Combining both routes gives the strongest result: full PVF reaches $31.34$ NFE, a further $27.6\%$ reduction over AR Fallback alone. This suggests that frequent local filling and selective planning are complementary: fallback supplies steady decoding progress, while planning adds extra gains when the low-confidence region contains high-value candidates. In other words, the result highlights the importance of exploiting high-value low-confidence positions when they are verifiable, which is precisely the role of PVF's Planning Route.

\section{Conclusion}

We introduced \textbf{Plan-Verify-Fill (PVF)}, a training-free decoding strategy for DLMs that uses limited verification budget to turn selected low-confidence candidates into reliable decoding progress. PVF preserves the high-confidence decoding backbone, verifies high-value planning drafts when informative uncertain positions are available, and falls back to conservative local filling otherwise. Experiments show consistent efficiency improvements while maintaining comparable accuracy, highlighting that low-confidence planning and local fill-in are complementary: effective DLM acceleration comes not from uniformly lowering confidence thresholds, but from spending verification on high-value uncertain positions when they can be safely accepted.

\section*{Limitations}

\paragraph{Scope.}
PVF is a training-free inference-time framework for adapting proposal--verification decoding to DLMs. Its central message is that masked decoding exposes a much larger space of candidate positions and generation orders than left-to-right decoding, while only a small number of additional drafts can be verified within the low-overhead regime. PVF should therefore be viewed as an initial step toward targeted draft selection, rather than a universal solution for all DLM decoding settings.

\paragraph{Model-internal verification.}
PVF relies on model-internal verification. Impact-set checks test whether an exploratory commitment remains compatible with the model's current high-confidence denoising trajectory, while local consistency checks play a similar role for AR-style fallback fills. These checks are lightweight and training-free, but they are not external correctness guarantees: a draft that passes verification is locally consistent with the model's own trajectory, not necessarily globally correct. Thus, PVF improves the inference-time accuracy--efficiency trade-off of an existing DLM, but does not replace stronger training-time alignment, external verifiers, or self-correction mechanisms.

\paragraph{Fixed calibration.}
The current implementation uses fixed calibration choices. The planning candidate set, confidence bands, and verification width are chosen before evaluation rather than adapted dynamically to each prompt, model state, or task instance. This keeps PVF simple and inexpensive, and every proposed commitment is still verified online before acceptance. Nevertheless, different model families, tokenizers, languages, or domains may benefit from different proposal policies. A natural extension is to learn or adapt the proposal distribution online, so that the limited verification budget is allocated to candidates with higher expected verified utility.

\paragraph{Systems dependence.}
Batched verification is not literally cost-free. The ``free-lunch'' region refers to a small-width systems regime in which evaluating a few additional branches adds little incremental wall-clock latency, often because GPU inference remains memory-bandwidth limited and accepted drafts amortize the verifier call. Outside this regime, larger verification widths can increase memory pressure and latency, while rejected candidates can turn parallelism into wasted computation. PVF deliberately keeps the verification width small, but realized wall-clock speedups may vary with accelerator type, inference backend, batching policy, memory layout, and serving load.

\paragraph{Evaluation coverage.}
Our experiments focus on a limited set of DLM backbones, benchmarks, block sizes, and generation lengths. While these settings cover representative reasoning and knowledge tasks, broader deployment may require retuning confidence bands, block configurations, and verification width for the target environment. We view this as part of the broader direction opened by PVF: efficient DLM decoding is not only about accepting more tokens, but about deciding which small set of exploratory commitments is worth verifying.

\bibliography{main}

\appendix

\section{Appendix}
\label{sec:appendix}

\subsection{Additional Background of {Semi-Autoregressive Blockwise Decoding}}\label{sec:block}

Semi-autoregressive diffusion adopts a hybrid structure combining \textit{inter-block} autoregression with \textit{intra-block} parallel diffusion. Widely adopted in modern DLMs~\cite{arriola2025block, nie2025large, sahoo2024simple, wu2025fast2,wu2025fast}, this strategy restricts attention to local regions, thereby reducing inference cost while preserving the local context. First partition the fixed canvas length $L$ into $B$ blocks, writing  
$\mathbf{y}_t = [y^{1:L/b}_t , y^{L/b+1:2(L/b)}_t , \ldots, y^{L-b+1:L}_t ]. $
The $k$-th block encompasses the token segment spanning from index $(k-1)\frac{L}{B}+1$ to $k\frac{L}{B}$. During inference, the decoding process adheres to a block-causal constraint: the state $\mathbf{y}_t$ can commit tokens in the $(k+1)$-th block only if the $k$-th block is fully resolved; otherwise, all subsequent blocks $j > k$ must remain masked. In this paper, we follow the framework of blockwise decoding, and define $\mathcal{B}_t$ to represent the set of indices for the current {working block} where \texttt{[MASK]} tokens are eligible for replacement at step $t$. 

\subsection{Planning Token Set Construction}
\label{sec:planning tokens}

\paragraph{Overview.}
Section~\ref{sec:planning-token-intuition} defines two event-level quantities for a candidate planning draft: verified compatibility $A_t^j$ and verified information gain $\mathrm{VIG}_t^j$. Here we describe how these event-level scores are aggregated into the planning-token vocabulary $P^{\mathrm{plan}}$. The construction is performed offline on non-evaluation data, and the resulting vocabulary is frozen before evaluation. Importantly, $P^{\mathrm{plan}}$ is only an eligibility filter for proposing candidates. It does not approve any commitment by itself, since every proposed planning draft is still verified online by PVF.

\paragraph{Calibration trajectory.}
For each task family, we run the target DLM on a calibration set $\mathcal{D}_{\mathrm{cal}}$ using the same confidence-based base decoding rule as in Section~\ref{sec:planning-token-intuition}. At each calibration step $t$, we form the base set $S_t^{\mathrm{base}}$, the base branch $\mathbf{z}_t^{\mathrm{base}}$, and the impact set $S_t^{\mathrm{impact}}$. We then inspect only non-base masked positions whose top-token confidence lies in a fixed calibration band:
\[
\mathcal{C}_t^{\mathrm{cal}}
=
\left\{
i\in M_t\setminus S_t^{\mathrm{base}}
\,\middle|\,
\tau_{\mathrm{cal}}^{\ell}
\le
p_t^i
<
\tau_{\mathrm{cal}}^u
\right\}.
\]
This band focuses calibration on the uncertain region where planning is potentially useful, while avoiding candidates that are either too unstable or already close to the high-confidence base rule.

For each candidate position $i\in \mathcal{C}_t^{\mathrm{cal}}$, we let $v=\hat y_t^i$ be its predicted vocabulary token and temporarily index it as candidate $j$ with $i_{(j)}=i$. We then construct the planning branch $\mathbf{z}_t^{\mathrm{plan},j}$ as in Eq.~\ref{eq:plan_branch}, compute the verified-compatibility outcome $A_t^j$ as in Eq.~\ref{eq:at}, and compute the event-level verified information gain $\mathrm{VIG}_t^j$ as defined in Section~\ref{sec:planning-token-intuition}. These branches are used only for measurement and are not committed to the calibration trajectory.

\paragraph{Token-level aggregation.}
Let $\mathcal{E}_v$ denote all calibration candidate events whose predicted vocabulary token is $v$:
\[
\mathcal{E}_v
=
\left\{
(t,j)
\,\middle|\,
\hat y_t^{i_{(j)}}=v,
\ 
i_{(j)}\in \mathcal{C}_t^{\mathrm{cal}}
\right\}.
\]
For each vocabulary token $v$, we compute four statistics:
\[
\begin{aligned}
n_v
&=
|\mathcal{E}_v|,\\
m_v
&=
\sum_{(t,j)\in\mathcal{E}_v}
\mathbf{1}\!\left[
|S_t^{\mathrm{impact}}|>0
\right],\\
r_v^{\mathrm{ver}}
&=
\frac{
\sum_{(t,j)\in\mathcal{E}_v}
A_t^j
}{
m_v
},\\
\widehat{\mathrm{VIG}}(v)
&=
\frac{1}{n_v}
\sum_{(t,j)\in\mathcal{E}_v}
\mathrm{VIG}_t^j .
\end{aligned}
\]
Here, $n_v$ measures support, $m_v$ counts non-vacuous verification opportunities, $r_v^{\mathrm{ver}}$ measures how often the token passes verified compatibility, and $\widehat{\mathrm{VIG}}(v)$ measures its average verified information gain. If $m_v=0$, we set $r_v^{\mathrm{ver}}=0$. Since $\mathrm{VIG}_t^j$ already includes the factor $A_t^j$, failed or vacuous candidates contribute zero information gain.

\begin{table*}[h]
\centering
\caption{Comparison of Planning versus Random token strategies in low-confidence regimes. The average confidence (\textit{Avg Token Conf}) of the selected tokens is maintained at comparable levels to ensure a controlled comparison.}
\begin{tabular}{ccccccc}
\toprule
\textbf{Dataset} &
$\bm{\tau}^{l}$ &
$\bm{\tau}^{u}$ &
\textbf{Method} &
\textbf{Acc.} &
\textbf{NFE} &
\textbf{Avg. Token Conf.} \\
\midrule
\multirow{9}{*}{GSM8K} & \multirow{2}{*}{0.10} & \multirow{2}{*}{0.60} & Planning & 0.73 & 50.67 & 0.38 \\
 &  &  & Random & 0.70 & 53.97 & 0.38 \\
\cmidrule(lr){2-7}
 & \multirow{2}{*}{0.15} & \multirow{2}{*}{0.60} & Planning & 0.74 & 49.63 & 0.40 \\
 &  &  & Random & 0.71 & 51.97 & 0.39 \\
\cmidrule(lr){2-7}
 & \multirow{2}{*}{0.20} & \multirow{2}{*}{0.60} & Planning & 0.75 & 49.78 & 0.42 \\
 &  &  & Random & 0.72 & 52.04 & 0.41 \\
\cmidrule(lr){2-7}
 & \multirow{2}{*}{0.25} & \multirow{2}{*}{0.60} & Planning & 0.76 & 50.55 & 0.43 \\
 &  &  & Random & 0.74 & 51.89 & 0.44 \\
\midrule
\multirow{9}{*}{HumanEval} & \multirow{2}{*}{0.10} & \multirow{2}{*}{0.60} & Planning & 0.24 & 66.38 & 0.37 \\
 &  &  & Random & 0.18 & 78.57 & 0.36 \\
\cmidrule(lr){2-7}
 & \multirow{2}{*}{0.15} & \multirow{2}{*}{0.60} & Planning & 0.31 & 67.54 & 0.39 \\
 &  &  & Random & 0.24 & 68.74 & 0.38 \\
\cmidrule(lr){2-7}
 & \multirow{2}{*}{0.20} & \multirow{2}{*}{0.60} & Planning & 0.30 & 66.99 & 0.41 \\
 &  &  & Random & 0.24 & 70.87 & 0.40 \\
\cmidrule(lr){2-7}
 & \multirow{2}{*}{0.25} & \multirow{2}{*}{0.60} & Planning & 0.30 & 68.90 & 0.43 \\
 &  &  & Random & 0.26 & 70.39 & 0.43 \\
\bottomrule
\end{tabular}
\end{table*}\label{table:sec3 ablation}

\begin{figure*}[t]
    \centering
    \includegraphics[width=0.24\textwidth]{ 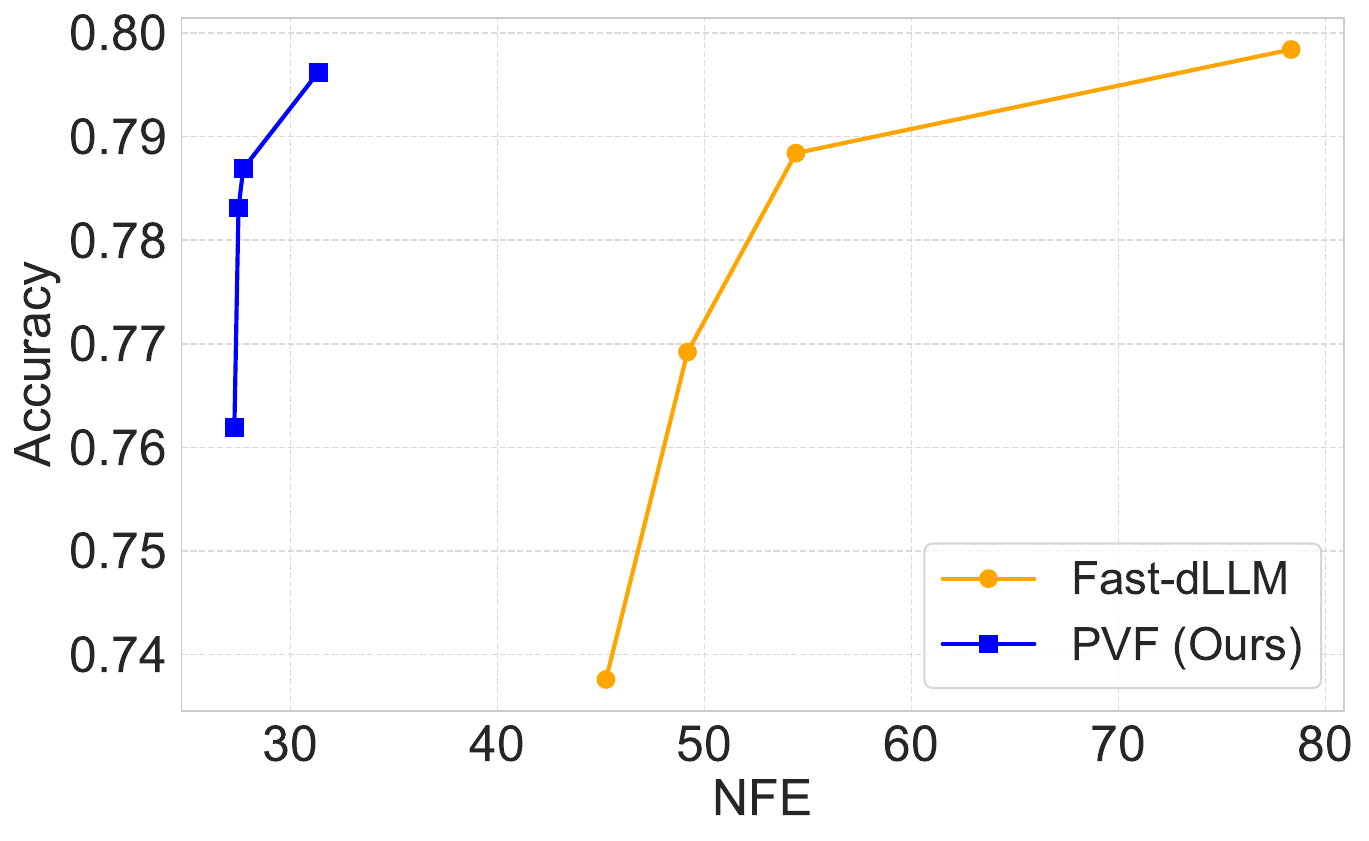}
    \includegraphics[width=0.24\textwidth]{ 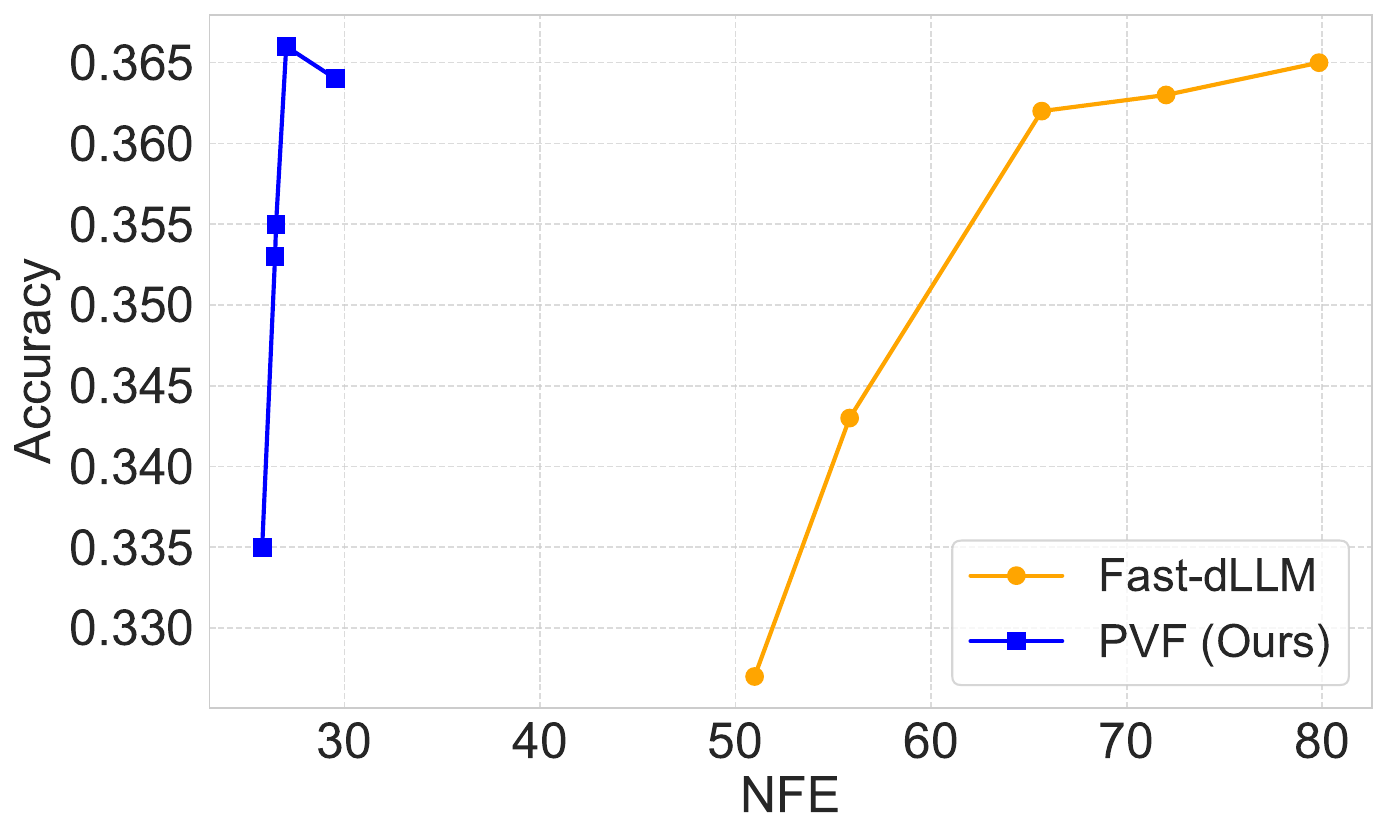}
    \includegraphics[width=0.24\textwidth]{ 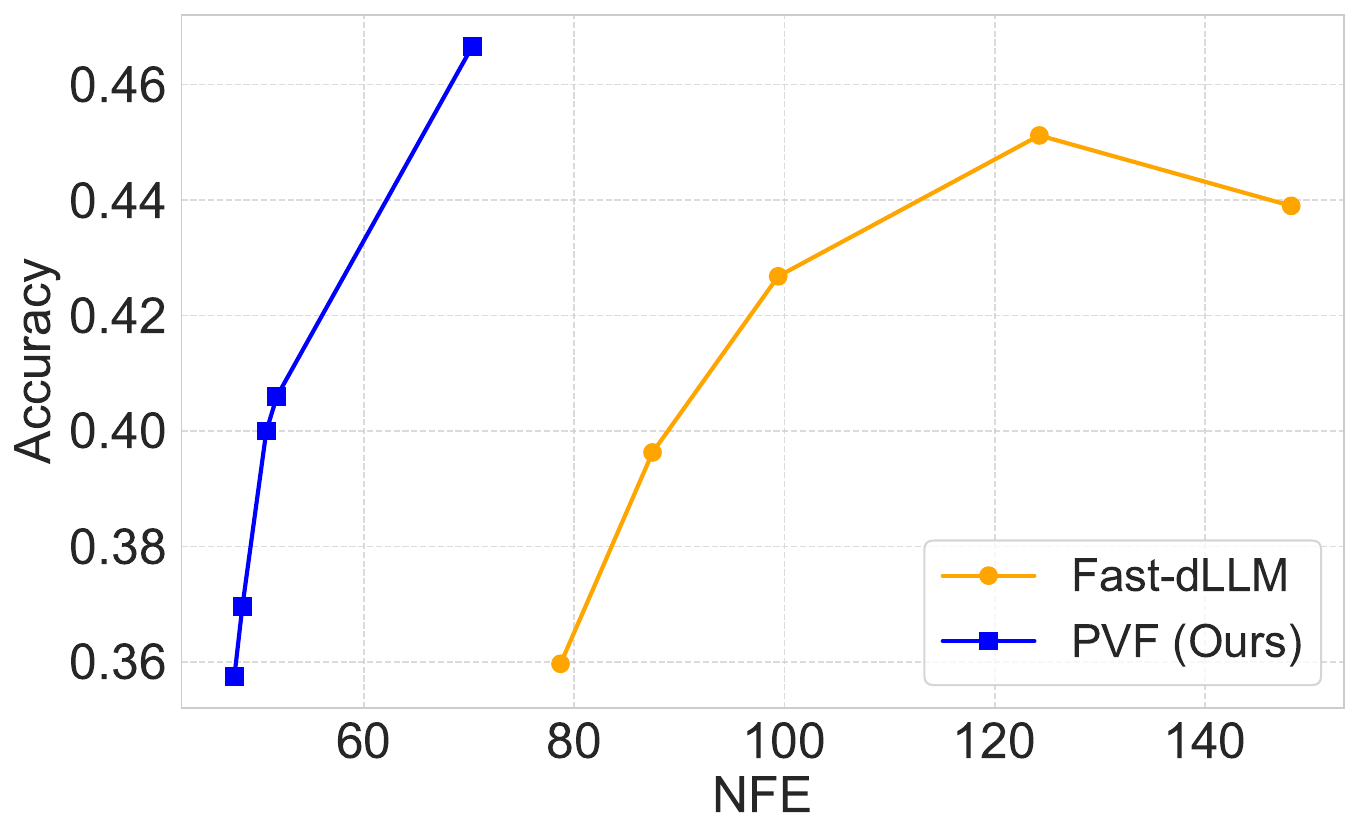}
    \includegraphics[width=0.24\textwidth]{ 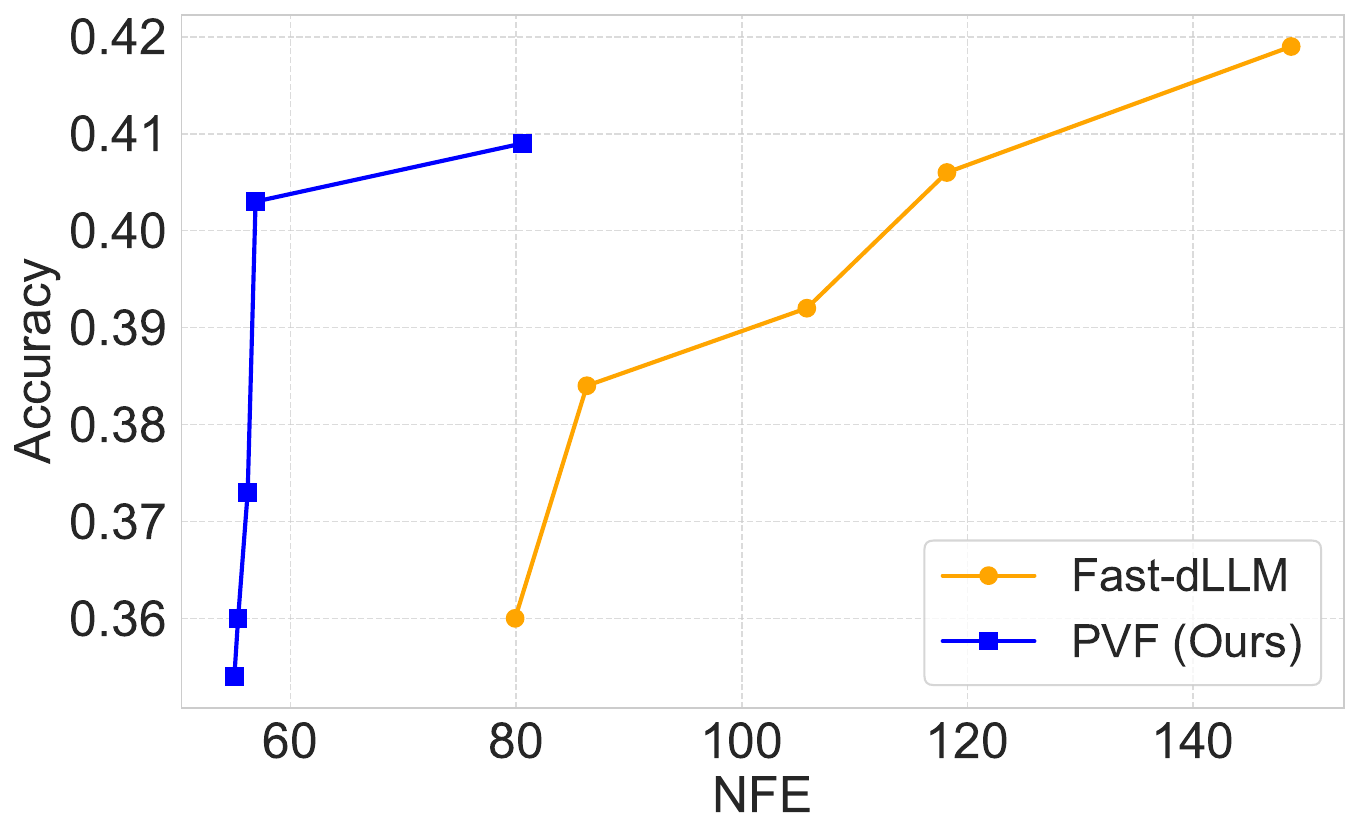}
    \caption{Pareto frontiers for PVF (blue) and Fast-dLLM (orange) on the GSM8K, MMLU-Pro, HumanEval, and Math datasets (from left to right) with LLaDA-8B-Instruct as the base model.}
    \label{fig:pareto}
    \vspace{-0.2cm}
\end{figure*}

\paragraph{Filtering rule.}
We retain a vocabulary token only if it has enough support, enough non-vacuous verification evidence, high verified-compatibility rate, and sufficient average verified information gain:
\[
P^{\mathrm{plan}}
=
\left\{
v\in V
\,\middle|\,
\begin{aligned}
&n_v\ge n_{\min},
\quad
m_v\ge m_{\min},\\
&r_v^{\mathrm{ver}}\ge \rho_{\mathrm{ver}},
\quad
\widehat{\mathrm{VIG}}(v)\ge \gamma_{\mathrm{vig}}
\end{aligned}
\right\}.
\]
This rule selects tokens by measured behavior under the target model, rather than by manual semantic labels.

\paragraph{Calibration data and thresholds.}
We construct $P^{\mathrm{plan}}$ separately for each task family. For mathematical reasoning tasks, we sample from the MATH training split. For code-generation tasks, we sample from the OpenCoder training corpus. For MMLU-style knowledge tasks, we sample from the \texttt{auxiliary\_train} split of CAIS MMLU. In all cases, we use $N_{\mathrm{cal}}=1000$ calibration examples, and labels are not used during calibration.

Unless otherwise stated, we use the calibration band
\[
[\tau_{\mathrm{cal}}^{\ell},\tau_{\mathrm{cal}}^u)
=
[0.15,0.25),
\]
and the filtering thresholds
\[
n_{\min}=50,
m_{\min}=20,
\rho_{\mathrm{ver}}=0.9,
\gamma_{\mathrm{vig}}=0.03.
\]
The selected vocabulary is fixed before evaluation and is used only to restrict the planning proposal space. Online verified compatibility remains the final acceptance criterion for every planning commitment.

\subsection{Additional Experiments}
\label{sec:add_exp}
\subsubsection{\emph{Planning} Tokens Beyond High-Confidence Decoding}\label{sec:complementary of sec3 ablation}
We evaluate two exploration variants---\emph{Random} and \emph{Planning}---under the same Semi-Autoregressive Blockwise Decoding framework introduced in Section~\ref{sec:block}. At each decoding iteration, we proceed as follows.
\begin{figure*}
    \centering    
    \includegraphics[width=0.9\linewidth]{ 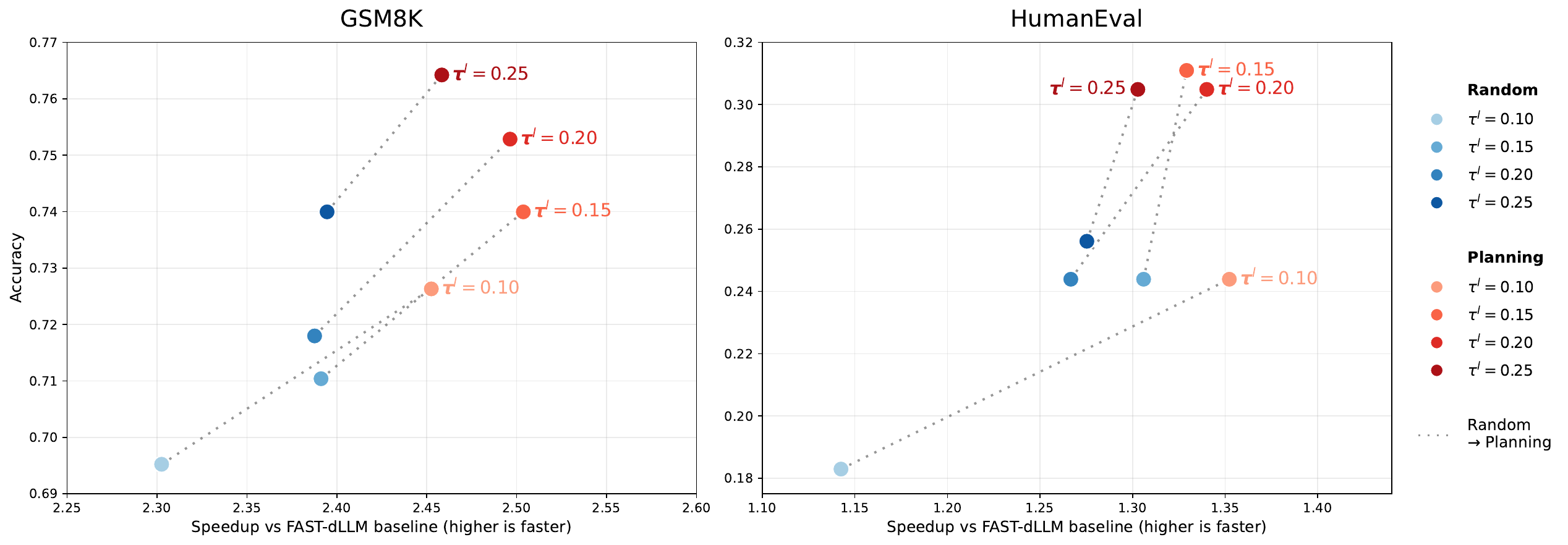}
    \vspace{-0.2cm}
\caption{Ablation on GSM8K and HumanEval comparing lower-confidence commits of planning tokens versus random tokens (i.e., without prioritizing planning tokens). Across confidence bins, prioritizing planning tokens consistently yields faster decoding and improved accuracy; points closer to the upper-right indicate better performance on both axes.}
\label{fig:token_ablation}
\end{figure*}

\begin{enumerate}
    \item \textbf{High-confidence commits.}
    We first commit all masked positions that satisfy the static Fast-dLLM unmasking rule \cite{wu2025fast} with threshold $\tau_{\text{high}} = 0.9$.

    \item \textbf{Low-confidence exploration within a confidence bin.}
    Let $[\tau^{l}, \tau^{u}]$ denote the active confidence range (bin). Among the remaining masked positions in the current active block whose top-1 posterior confidence lies in $[\tau^{l}, \tau^{u}]$, we commit exactly one additional token using one of the following strategies:
    \begin{enumerate}
        \item \textbf{Random.}
        Uniformly sample one eligible masked position (within the block) and commit its top-1 token.

        \item \textbf{Planning.}
        We form the subset of eligible masked positions whose top-1 token is in $\mathcal{P}$ and whose confidence lies in $[\tau^{l}, \tau^{u}]$. If this subset is non-empty, we uniformly sample one element from it and commit its top-1 token. If it is empty, we fall back to the same Random rule above (sampling from all eligible positions in $[\tau^{l}, \tau^{u}]$).
    \end{enumerate}
\end{enumerate}

The fallback in the Planning variant is important for a fair comparison: since ${P}^{\text{plan}}$ is deliberately small, the planning-eligible subset is sometimes empty (see Figure~\ref{fig:plan_rate}). Without a fallback, the Planning variant would make fewer low-confidence commits, confounding accuracy and runtime with a different number of exploration steps. By enforcing a matched exploration rate via fallback, improvements can be attributed to \emph{which} low-confidence tokens are committed rather than \emph{how many} are committed.
Empirically, we adopt the static variant of Fast-DLLM \citep{wu2025fast} as the baseline. At each decoding step, in addition to the tokens selected by Fast-DLLM, we randomly commit one extra token whose confidence lies within a prescribed low-confidence interval $[\tau^l,\tau^u]$. In Figure~\ref{fig:token_ablation}, we fix $\tau^u=0.6$ for simplicity. The results show that unrestricted low-confidence commitments, labeled \emph{Random} in Figure~\ref{fig:token_ablation}, cause substantial performance degradation, whereas prioritizing our content-neutral planning token set $P^{\mathrm{plan}}$ for such commitments, labeled \emph{Planning} in Figure~\ref{fig:token_ablation}, markedly reduces the accuracy loss while also improving efficiency.

Additionally, Table~\ref{table:sec3 ablation} reports the average confidence of the extra lower-confidence tokens committed under both the \textbf{Random} and \textbf{Planning} strategies. Across all datasets, the average values differ by at most 0.01, indicating that the gains in efficiency and accuracy are driven by prioritizing commits to planning tokens, rather than due to a statistical bias whereby planning tokens exhibit systematically higher posterior confidence within the same confidence band.

\paragraph{Speedup metric.}
We quantify decoding speed by the number of function evaluations (NFE), i.e., the total number of forward passes. For each setting, we report speedup as
\[
\text{Speedup} \;=\; \frac{\mathrm{NFE}_{\text{Fast-dLLM}}}{\mathrm{NFE}_{\text{(Random/Planning)}}},
\]
so that larger values indicate fewer forward passes and faster decoding.

\subsubsection{Pareto Frontier}
Figure~\ref{fig:pareto} plots the accuracy--NFE trade-off of PVF and Fast-dLLM under a sweep of the shared high-confidence threshold $\tau_{\mathrm{high}}$. For a fair comparison, we keep PVF-specific hyperparameters fixed and vary only the shared threshold used by both methods. Across four benchmark datasets, PVF consistently shifts the frontier toward lower NFE at comparable accuracy. This means that for the same accuracy target, PVF generally requires substantially fewer forward passes than Fast-dLLM. Conversely, for a fixed NFE budget, PVF can often operate at a higher-accuracy point. The advantage is most visible near the high-accuracy regime, where Fast-dLLM must use a conservative threshold and therefore commits fewer tokens per denoising step, while PVF can still make additional verified progress through planning and fallback routes. These results indicate that PVF improves the accuracy--efficiency frontier rather than merely tuning the confidence threshold more aggressively. 

\begin{table*}[h]
\centering
\caption{Comparison of PVF and Fast-dLLM performance across different Generation Lengths and Block Sizes on GSM8K for LLaDA-8B-Instruct.}
\begin{tabular}{cc c c c c}
\toprule
\multirow{2}{*}{\textbf{Gen Length}} & \multirow{2}{*}{\textbf{Block Size}} & \multicolumn{2}{c}{\textbf{PVF}} & \multicolumn{2}{c}{\textbf{Fast-dLLM}} \\
\cmidrule(lr){3-4} \cmidrule(lr){5-6}
 & & \textbf{Accuracy} & \textbf{NFE} & \textbf{Accuracy} & \textbf{NFE} \\
\midrule
\multirow{3}{*}{256} & 16 & 76.46 & 37.76 & 76.30 & 77.73 \\
                     & 32 & 75.92 & 26.75 & 75.76 & 71.23 \\
                     & 64 & 75.84 & 24.52 & 75.69 & 67.85 \\
\midrule
\multirow{3}{*}{512} & 16 & 79.53 & 58.34 & 80.07 & 110.32 \\
                     & 32 & 79.92 & 41.55 & 79.69 & 96.86 \\
                     & 64 & 79.62 & 31.34 & 79.62 & 89.77 \\
\bottomrule
\end{tabular}
\label{tab:gsm8k_comparison}
\end{table*}

\subsubsection{Sensitivity Analysis: Generation Length and Block Size}
To evaluate the robustness of our approach, we benchmark PVF against the Fast-dLLM baseline across varying maximum generation lengths ($L \in \{256, 512\}$) and block sizes ($k \in \{16, 32, 64\}$) on the GSM8K dataset. As shown in Table~\ref{tab:gsm8k_comparison}, PVF consistently demonstrates a substantial improvement in algorithmic efficiency, achieving significantly lower NFE than Fast-dLLM across all configurations while maintaining near-identical accuracy. Furthermore, we observe a positive correlation between block size and efficiency gains: as the block size increases, the reduction in NFE becomes more pronounced. This trend validates the efficacy of our planning mechanism, which leverages larger blocks to secure longer verified decoding trajectories, thereby minimizing the number of required forward passes more effectively.

\subsubsection{Sensitivity Analysis: Planning token thresholds $\tau_{\text{plan}}^l$, $\tau_{\text{plan}}^u$ }

We further investigate the sensitivity of PVF to the planning-token thresholds, specifically the lower bound $\tau_{\text{plan}}^l$ and upper bound $\tau_{\text{plan}}^u$. To ensure a fair comparison, we fix all other hyperparameters to match those used in Table~\ref{tab:main_results_acc_nfe} for both datasets. Across the range of values we test, accuracy remains largely stable, indicating that the planning-token procedure is reasonably reliable. We nonetheless observe a mild trade-off between algorithmic efficiency and generation quality: lowering both threshold bounds consistently reduces NFE (improving efficiency) but leads to a corresponding minor decline in accuracy. This behavior occurs because lowering the thresholds effectively forces the model to commit to planning tokens with lower confidence. While this aggressive commitment typically provides a stronger contextual signal that boosts the confidence of subsequent tokens, which accelerates the decoding process, it simultaneously increases the chance of making mistakes.

\begin{figure}[H]
    \centering
\includegraphics[width=0.4\linewidth]{ 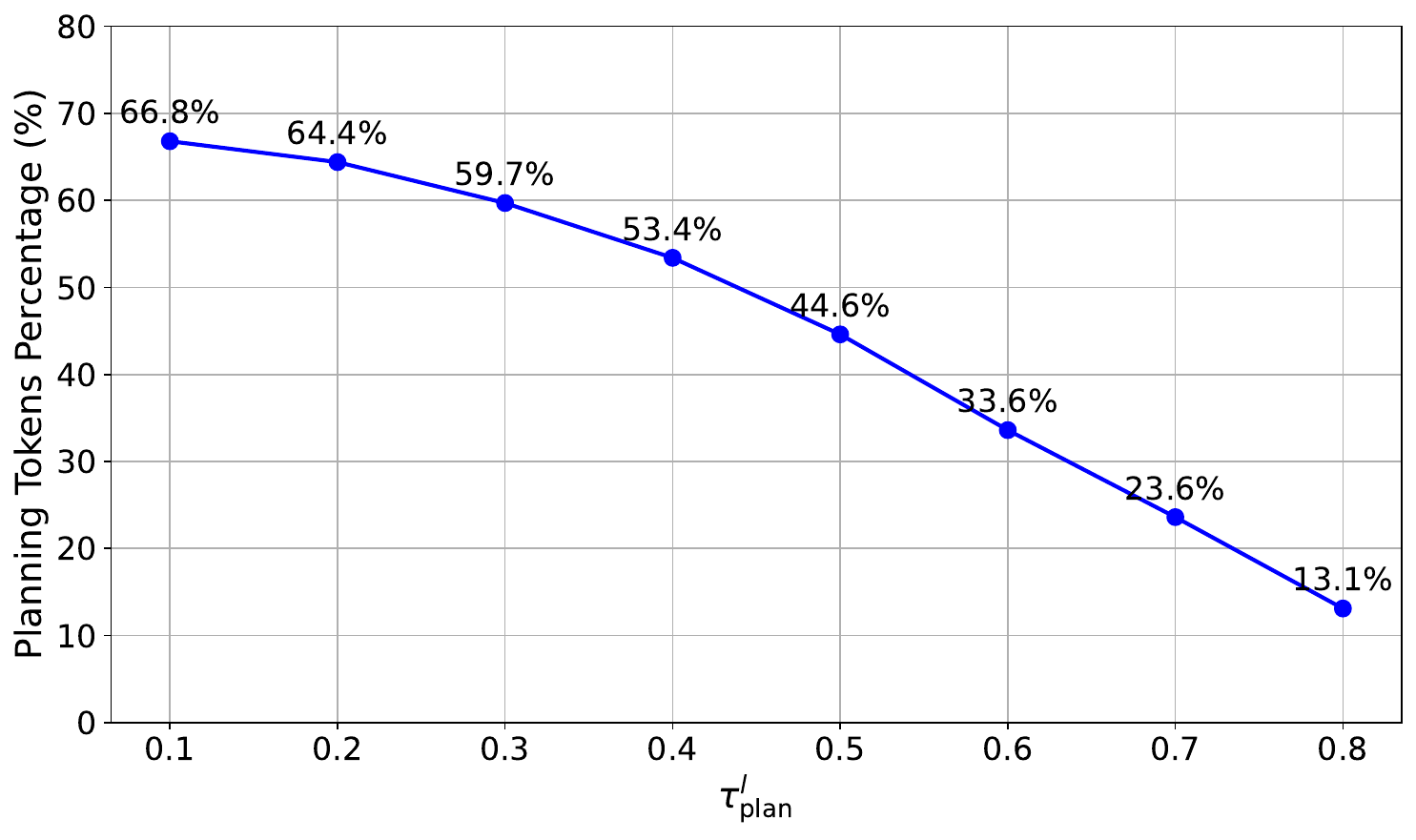}
    \caption{Impact of the planning token threshold lower bound ($\tau_{\text{plan}}^l$) on the percentage of selected planning tokens.}
    \label{fig:plan_rate}
\end{figure}

We further examine the percentage of planning tokens relative to the total token count (defined as the sum of planning and AR fallback tokens) under varying lower bound thresholds $\tau_{\text{plan}}^l$, while fixing the upper bound at $\tau_{\text{plan}}^u=0.9$. Figure~\ref{fig:plan_rate} illustrates this relationship, showing that a lower threshold results in a higher proportion of selected planning tokens, which in turn typically leads to better algorithmic efficiency.

\subsection{Ablation Study: Sensitivity to Batch Size}\label{sec:Ablation Study: Sensitivity to Batch Size}
\begin{figure}[H]
    \centering
    \includegraphics[width=0.4\linewidth]{ 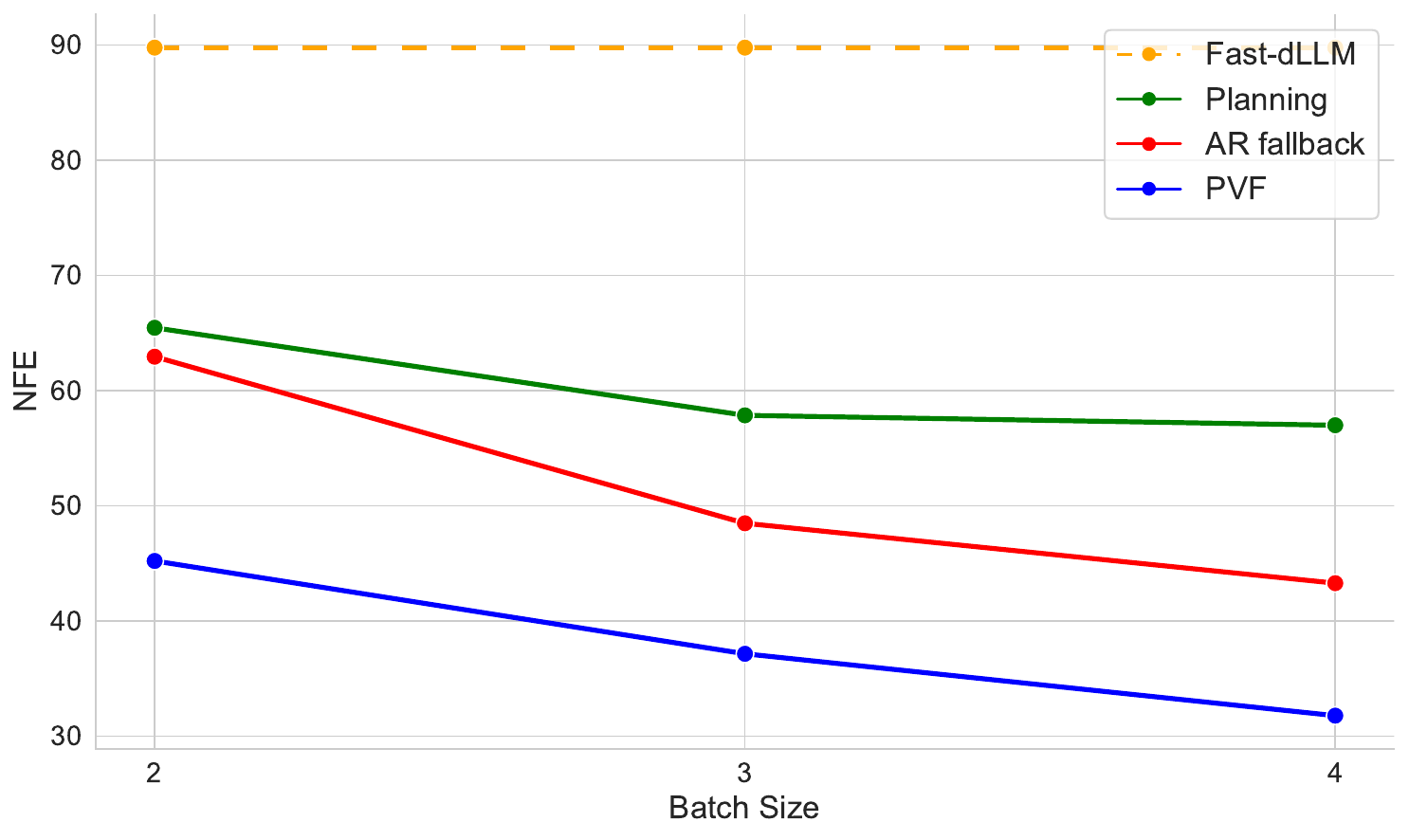}
    \caption{Ablation study on the full GSM8k dataset evaluating the impact of PVF components across varying batch sizes in the ``free lunch'' regime. All results maintain lossless accuracy compared to static decoding.}
    \label{fig:ablation_batch}
\end{figure}

To assess whether our algorithm retains efficiency even with more constrained batch sizes, we evaluate the NFE of our PVF method and its individual components as the batch size varies, ensuring that accuracy remains lossless relative to static decoding. In Figure~\ref{fig:ablation_batch}, we observe that the planning stage delivers stable performance even at a batch size of $2$, whereas the AR fallback benefits more significantly from larger batches due to its cumulative nature. Notably, the full PVF method consistently outperforms both individual components across all settings, and demonstrates substantial efficiency gains compared to the baseline method even at a batch size of $2$.

\newpage
\onecolumn

\begin{figure}[H]
    \centering
\includegraphics[width=0.7\linewidth]{ 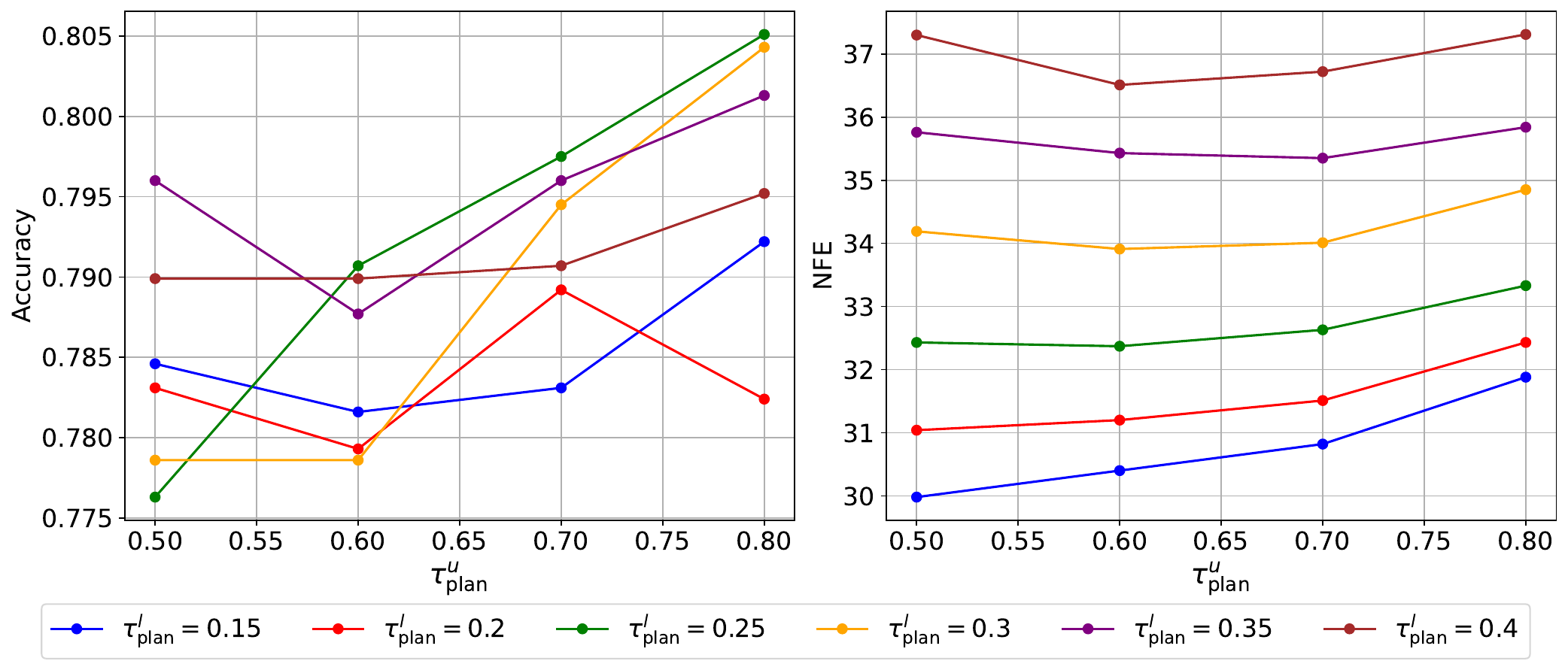}
    \caption{Sensitivity analysis of planning-token thresholds on GSM8K. Each curve corresponds to a fixed lower bound $\tau_{\text{plan}}^l$ and shows accuracy and NFE as $\tau_{\text{plan}}^u$ varies.}
    \label{fig:sens_thresh}
\end{figure}

\begin{figure}[H]
    \centering
\includegraphics[width=0.7\linewidth]{ 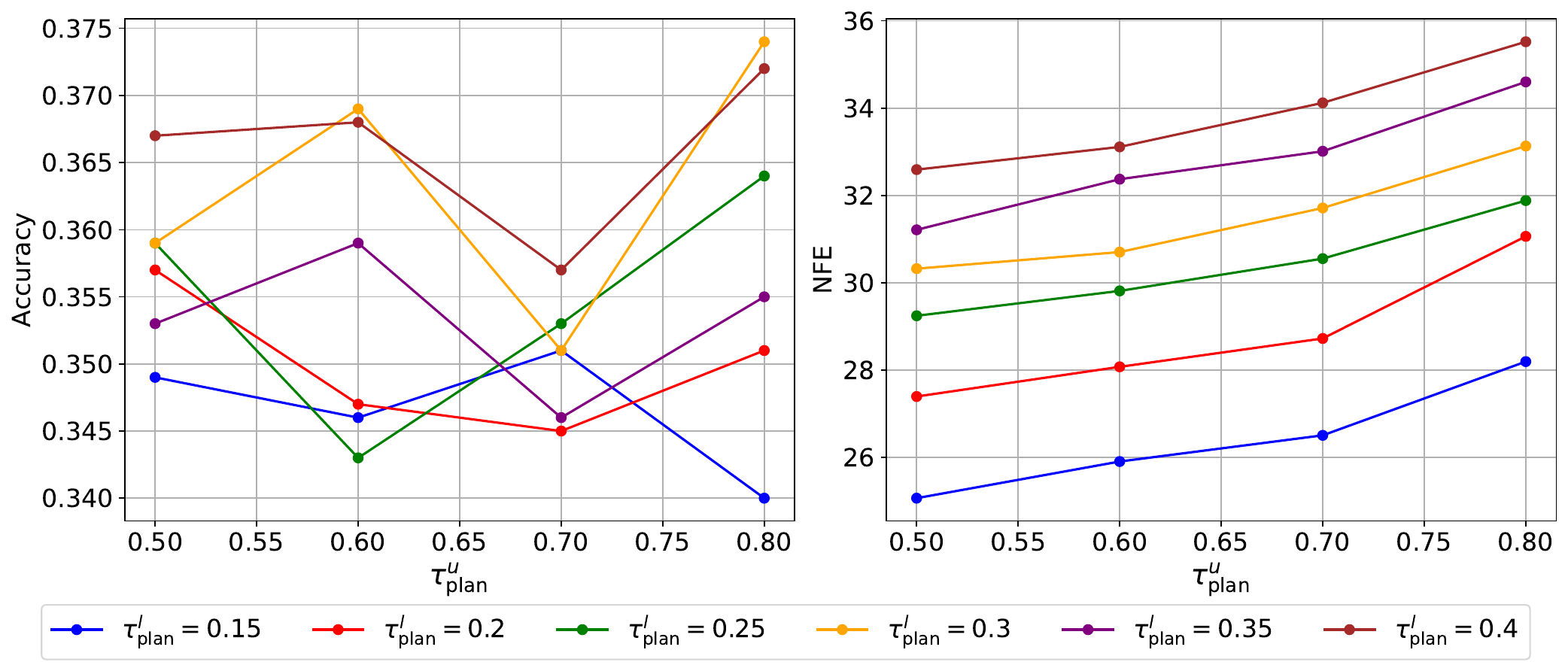}
    \caption{Sensitivity analysis of planning-token thresholds on MMLU-Pro. Each curve corresponds to a fixed lower bound $\tau_{\text{plan}}^l$ and shows accuracy and NFE as $\tau_{\text{plan}}^u$ varies.}
    \label{fig:sens_thresh}
\end{figure}

\newpage
\subsection{Additional Details of PVF}

\begin{figure*}[t]
    \centering
    \includegraphics[width=0.7\linewidth]{ 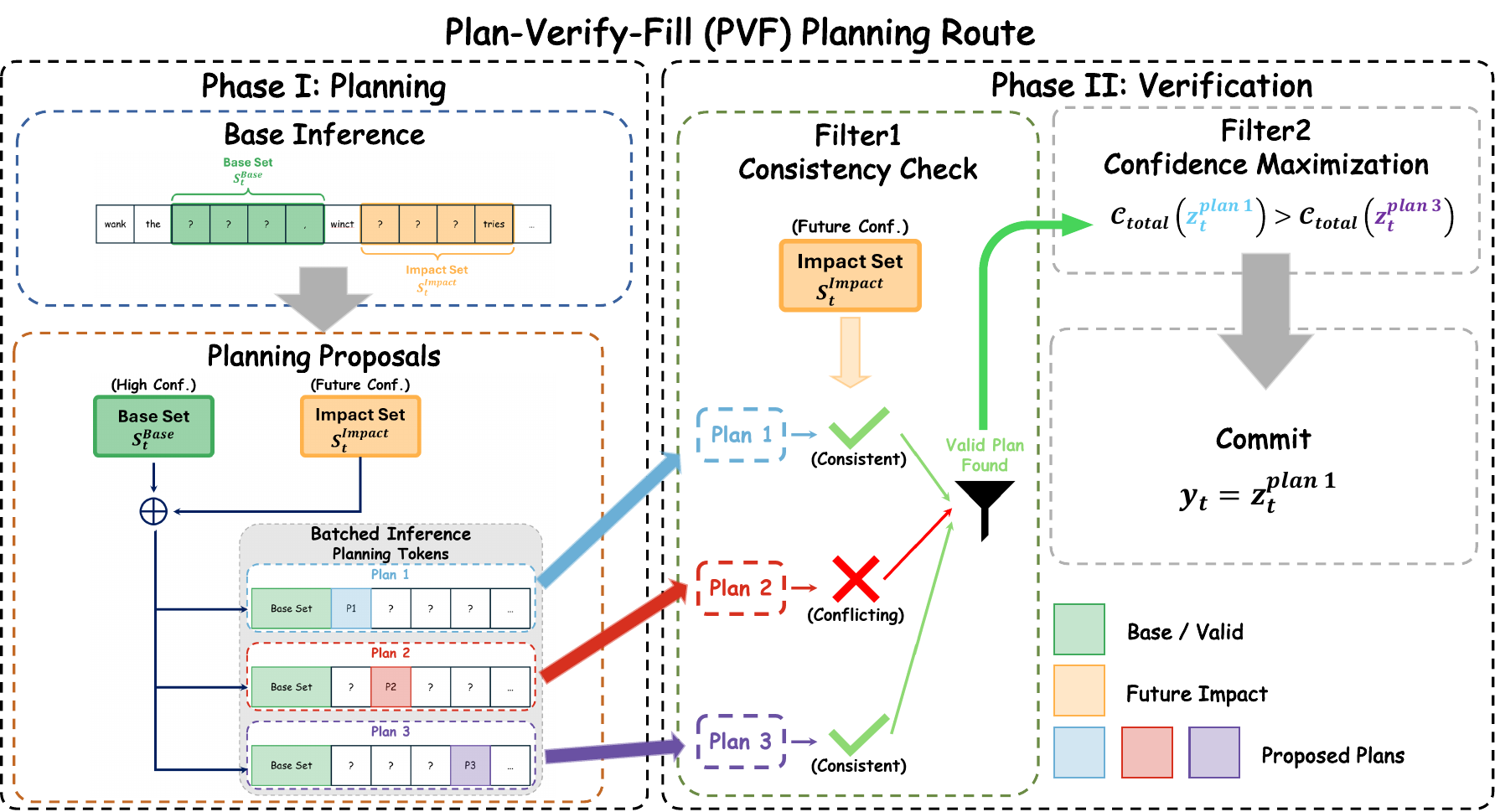}
    \includegraphics[width=0.7\linewidth]{ 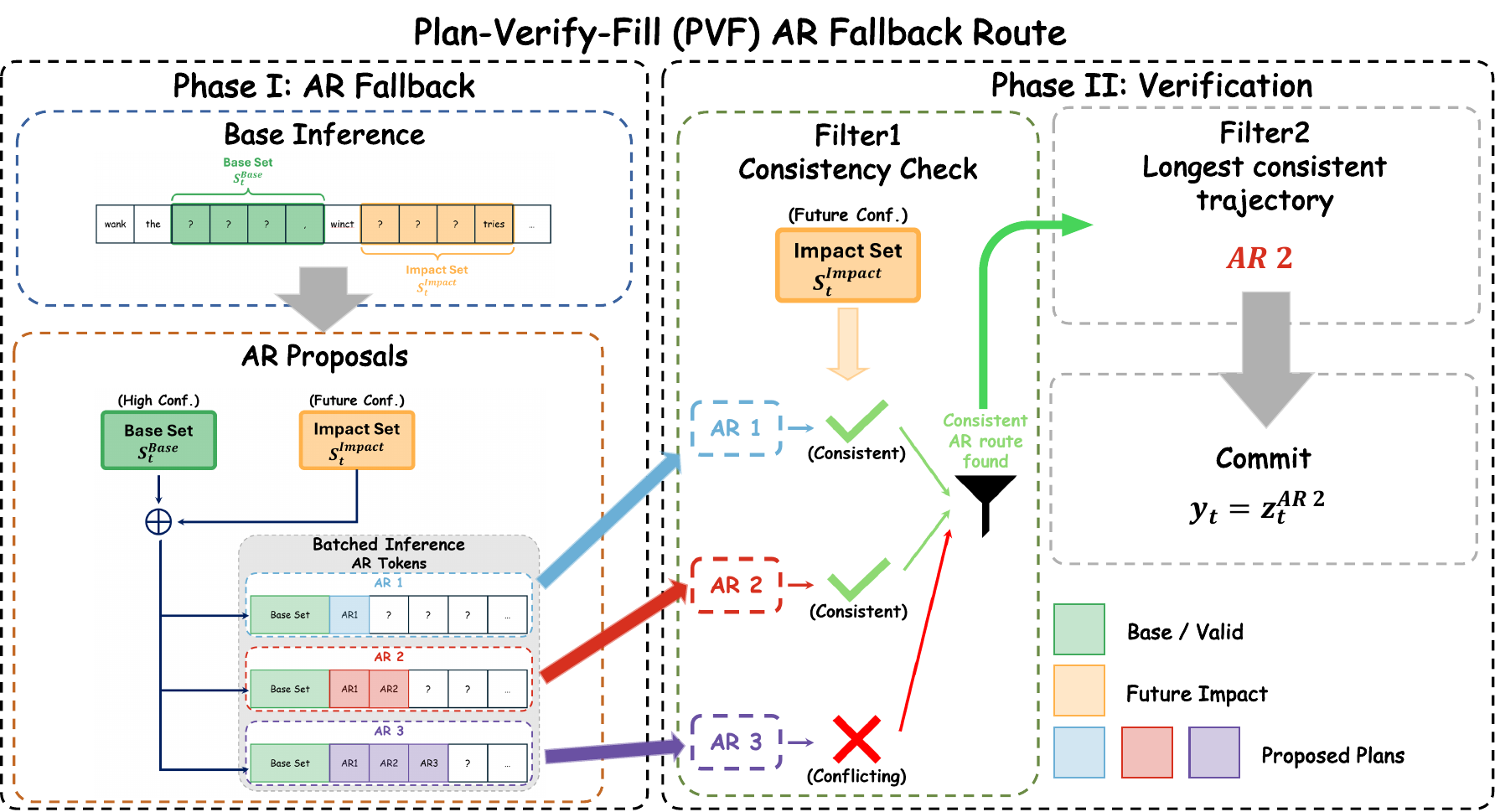}
\caption{Overview of the \textbf{Planning Route} and the \textbf{AR Fallback Route}.}
\label{fig:pvf_planning}
\vspace{-0.3cm}
\end{figure*}

\subsubsection{Planning-Guided Early Cross-Block Revelation}
While retaining block-wise efficiency, PVF discards the rigid requirement of strictly sequential convergence.
Our motivation is structural: as the active block enters a \textit{saturated regime} of mask scarcity, the constricted search space limits the planner's optimization capacity. To counteract this, we adopt a \textit{dynamic expansion strategy} that reveals the subsequent block preemptively when the current block's sparsity hinders effective planning.

Consistent with recent approaches that relax rigid block boundaries~\cite{wang2025diffusion, fu2025bits}, we allow for early cross-block revelation to facilitate information flow. However, to mitigate the accuracy degradation often observed with aggressive expansion~\cite{wang2025diffusion}, this strategy is implemented conservatively. Expansion is only triggered when the current block enters a saturated regime of mask scarcity, ensuring that the early decoding mechanism serves solely to maintain a valid planning horizon.

Formally, the block size is denoted as $S = L/B$ . We identify the index of the current active block $k_t$ as the first block containing masked tokens. Define the index set of masked tokens in the current active block $\mathcal{M}_{t}^{\text{active}} = \{i|y_{t-1}^i=\texttt{[MASK]}\}\bigcap \{S(k_t-1)+1,\ldots, Sk_t\}$. We define the working index set $\mathcal{B}_t$ by checking if the remaining masks in the current block fall below a sparsity threshold $N_{\text{s}}$:

\begin{align}\label{eqn:block_expansion}
    \mathcal{B}_t = \begin{cases} 
        \mathcal{M}_{t}^{\text{active}}\bigcup\mathcal{I}_{k_t+1} , & \text{if } \left| \mathcal{M}_{t}^{\text{active}} \right| \leq N_{\text{s}} \\[8pt]
       \mathcal{M}_{t}^{\text{active}}, & \text{otherwise}
    \end{cases}
\end{align}
where $\mathcal{I}_{k_t+1} = \{k_tS + 1, \dots, (k_t+1)S\}$ represents the full index range of the next block. This condition ensures that the planner always operates on a sufficiently broad horizon, seamlessly transitioning focus to $k+1$ as block $k_t$ stabilizes.

\subsection{Additional Experimental Details}\label{sec:Additional Experimental Details}
\paragraph{Experimental details.}
Across all datasets, we use a fixed maximum generation length of $L=512$ and a block size of $B=64$.
Decoding is training-free (no finetuning, no additional supervision, and no external verifier).
All experiments are conducted on a single NVIDIA H200 GPU.

\paragraph{Implementation and evaluation.}
All methods share the same prompt templates per dataset, the same stopping criterion, and the same maximum generation length.
For HumanEval, we follow the standard unit-test protocol and report pass@1.
For MATH and GSM8K, we use exact-match evaluation on the final answer.
For MMLU-Pro, we report multiple-choice accuracy.

\paragraph{Hyper-parameters.}
Table~\ref{tab:hyperparams} summarizes the hyperparameters used across benchmarks.

\begin{table}[H]
\centering
\small
\setlength{\tabcolsep}{10pt}
\renewcommand{\arraystretch}{1.15}
\begin{tabular}{p{1.7cm}|p{2cm}|p{2.8cm}|p{2cm}|p{2.5cm}}
\toprule
\textbf{Benchmark} &
\textbf{Base Budget} $(L,B)$ &
\textbf{Planning Thresholds} $(\tau^l_{\text{plan}},\tau^{u}_{\text{plan}})$ &
\textbf{AR Fallback Threshold $\tau_{\text{AR}}^{l}$} & \textbf{Early Cross-block Threshold $N_s$}\\
\midrule
MMLU-Pro &
$L{=}512,\;B{=}64$ &
$(0.2,\;0.65)$ &
$0.1$ & 5\\

GSM8K &
$L{=}512,\;B{=}64$ &
$(0.2,\;0.65)$ &
$0.1$ & 5 \\

HumanEval &
$L{=}512,\;B{=}64$ &
$(0.8,\;0.9)$ &
$0.3$ & 0 \\

MATH &
$L{=}512,\;B{=}64$ &
$(0.8,\;0.9)$ &
$0.3$ & 0 \\
\bottomrule
\end{tabular}
\caption{Report of Hyperparameters Used}
\label{tab:hyperparams}
\end{table}
For the Pareto sweeps in Fig.~\ref{fig:pareto} on HumanEval and MATH, we additionally enforce a feasibility constraint that the
planning range does not overlap with the high-confidence region $(\tau_{\text{high}},1)$, since overlapping these zones would make the
roles of ``planning'' versus ``high-confidence'' commits ill-defined. Concretely, we set
$(\tau^l_{\text{plan}},\tau^u_{\text{plan}})=(0.4,0.6)$ for $\tau_{\text{high}}<0.9$.
\clearpage

\onecolumn
\subsection{Case study}
In this section, we present two example generations from LLaDA-8B-Instruct using Fast-dLLM and our PVF strategy. For PVF, we highlight tokens committed via the planning route in yellow and tokens committed via the AR fallback route in pink.

\begin{tcolorbox}[colback=white, colframe=gray!80, title=LLaDA Prediction of Fast-dLLM (89 steps), sharp corners=south,before skip=5pt]
\tokenH{To}\tokenH{ determine}\tokenH{ the}\tokenH{ total}\tokenH{ number}\tokenH{ of}\tokenH{ hours}\tokenH{ Meredith}\tokenH{ spent}\tokenH{ writing}\tokenH{ articles}\tokenH{ over}\tokenH{ the}\tokenH{ three}\tokenH{ days}\tokenH{,}\tokenH{ we}\tokenH{ need}\tokenH{ to}\tokenH{ follow}\tokenH{ these}\tokenH{ steps}\tokenH{:}
\\
\\
\tokenH{1}\tokenH{.}\tokenH{ Calculate}\tokenH{ the}\tokenH{ number}\tokenH{ of}\tokenH{ articles}\tokenH{ she}\tokenH{ wrote}\tokenH{ on}\tokenH{ Tuesday}\tokenH{.}
\\
\tokenH{2}\tokenH{.}\tokenH{ Calculate}\tokenH{ the}\tokenH{ number}\tokenH{ of}\tokenH{ articles}\tokenH{ she}\tokenH{ wrote}\tokenH{ on}\tokenH{ Wednesday}\tokenH{.}
\\
\tokenH{3}\tokenH{.}\tokenH{ Find}\tokenH{ the}\tokenH{ total}\tokenH{ number}\tokenH{ of}\tokenH{ articles}\tokenH{ she}\tokenH{ wrote}\tokenH{ over}\tokenH{ the}\tokenH{ three}\tokenH{ days}\tokenH{.}
\\
\tokenH{4}\tokenH{.}\tokenH{ Calculate}\tokenH{ the}\tokenH{ total}\tokenH{ number}\tokenH{ of}\tokenH{ hours}\tokenH{ she}\tokenH{ spent}\tokenH{ writing}\tokenH{ the}\tokenH{ articles}\tokenH{.}
\\
\\
\tokenH{**}\tokenH{Step}\tokenH{ }\tokenH{1}\tokenH{:}\tokenH{ Calculate}\tokenH{ the}\tokenH{ number}\tokenH{ of}\tokenH{ articles}\tokenH{ she}\tokenH{ wrote}\tokenH{ on}\tokenH{ Tuesday}\tokenH{.**}
\\
\\
\tokenH{Meredith}\tokenH{ wrote}\tokenH{ 5}\tokenH{ articles}\tokenH{ on}\tokenH{ Monday}\tokenH{.}\tokenH{ On}\tokenH{ Tuesday}\tokenH{,}\tokenH{ she}\tokenH{ wrote}\tokenH{ \textbackslash{}(}\tokenH{ \textbackslash{}}\tokenH{frac}\tokenH{\{}\tokenH{2}\tokenH{\}}\tokenH{\{}\tokenH{5}\tokenH{\}}\tokenH{ \textbackslash{})}\tokenH{ times}\tokenH{ more}\tokenH{ articles}\tokenH{ than}\tokenH{ on}\tokenH{ Monday}\tokenH{.}\tokenH{ Therefore}\tokenH{,}\tokenH{ the}\tokenH{ number}\tokenH{ of}\tokenH{ articles}\tokenH{ she}\tokenH{ wrote}\tokenH{ on}\tokenH{ Tuesday}\tokenH{ is}\tokenH{:}\\
\tokenH{\textbackslash{}[}
\\
\tokenH{ 5}\tokenH{ +}\tokenH{ \textbackslash{}}\tokenH{frac}\tokenH{\{}\tokenH{2}\tokenH{\}}\tokenH{\{}\tokenH{5}\tokenH{\}}\tokenH{ \textbackslash{}}\tokenH{times}\tokenH{ 5}\tokenH{ =}\tokenH{ 5}\tokenH{ +}\tokenH{ 2}\tokenH{ =}\tokenH{ 7}\\
\tokenH{\textbackslash{}}\tokenH{]}\tokenH{}
\\
\\
\tokenH{**}\tokenH{Step}\tokenH{ }\tokenH{2}\tokenH{:}\tokenH{ Calculate}\tokenH{ the}\tokenH{ number}\tokenH{ of}\tokenH{ articles}\tokenH{ she}\tokenH{ wrote}\tokenH{ on}\tokenH{ Wednesday}\tokenH{.**}
\\
\\
\tokenH{On}\tokenH{ Wednesday}\tokenH{,}\tokenH{ she}\tokenH{ wrote}\tokenH{ twice}\tokenH{ the}\tokenH{ number}\tokenH{ of}\tokenH{ articles}\tokenH{ she}\tokenH{ wrote}\tokenH{ on}\tokenH{ Tuesday}\tokenH{.}\tokenH{ Therefore}\tokenH{,}\tokenH{ the}\tokenH{ number}\tokenH{ of}\tokenH{ articles}\tokenH{ she}\tokenH{ wrote}\tokenH{ on}\tokenH{ Wednesday}\tokenH{ is}\tokenH{:}\\
\tokenH{\textbackslash{}[}\\
\tokenH{ 2}\tokenH{ \textbackslash{}}\tokenH{times}\tokenH{ 7}\tokenH{ =}\tokenH{ 1}\tokenH{4}\\
\tokenH{\textbackslash{}}\tokenH{]}\tokenH{}
\\
\\
\tokenH{**}\tokenH{Step}\tokenH{ }\tokenH{3}\tokenH{:}\tokenH{ Find}\tokenH{ the}\tokenH{ total}\tokenH{ number}\tokenH{ of}\tokenH{ articles}\tokenH{ she}\tokenH{ wrote}\tokenH{ over}\tokenH{ the}\tokenH{ three}\tokenH{ days}\tokenH{.**}
\\
\\
\tokenH{The}\tokenH{ total}\tokenH{ number}\tokenH{ of}\tokenH{ articles}\tokenH{ she}\tokenH{ wrote}\tokenH{ over}\tokenH{ the}\tokenH{ three}\tokenH{ days}\tokenH{ is}\tokenH{ the}\tokenH{ sum}\tokenH{ of}\tokenH{ the}\tokenH{ articles}\tokenH{ she}\tokenH{ wrote}\tokenH{ on}\tokenH{ Monday}\tokenH{,}\tokenH{ Tuesday}\tokenH{,}\tokenH{ and}\tokenH{ Wednesday}\tokenH{:}\\
\tokenH{\textbackslash{}[}\\
\tokenH{ 5}\tokenH{ +}\tokenH{ 7}\tokenH{ +}\tokenH{ 1}\tokenH{4}\tokenH{ =}\tokenH{ 2}\tokenH{6}\\
\tokenH{\textbackslash{}}\tokenH{]}\tokenH{}
\\
\\
\tokenH{**}\tokenH{Step}\tokenH{ }\tokenH{4}\tokenH{:}\tokenH{ Calculate}\tokenH{ the}\tokenH{ total}\tokenH{ number}\tokenH{ of}\tokenH{ hours}\tokenH{ she}\tokenH{ spent}\tokenH{ writing}\tokenH{ the}\tokenH{ articles}\tokenH{.**}
\\
\\
\tokenH{Each}\tokenH{ article}\tokenH{ takes}\tokenH{ an}\tokenH{ average}\tokenH{ of}\tokenH{ 4}\tokenH{ hours}\tokenH{ to}\tokenH{ research}\tokenH{ and}\tokenH{ write}\tokenH{.}\tokenH{ Therefore}\tokenH{,}\tokenH{ the}\tokenH{ total}\tokenH{ number}\tokenH{ of}\tokenH{ hours}\tokenH{ she}\tokenH{ spent}\tokenH{ writing}\tokenH{ the}\tokenH{ articles}\tokenH{ is}\tokenH{:}\\
\tokenH{\textbackslash{}[}\\
\tokenH{ 2}\tokenH{6}\tokenH{ \textbackslash{}}\tokenH{times}\tokenH{ 4}\tokenH{ =}\tokenH{ 1}\tokenH{0}\tokenH{4}
\\
\tokenH{\textbackslash{}}\tokenH{]}\tokenH{}
\\
\\
\tokenH{Thus}\tokenH{,}\tokenH{ the}\tokenH{ total}\tokenH{ number}\tokenH{ of}\tokenH{ hours}\tokenH{ she}\tokenH{ spent}\tokenH{ writing}\tokenH{ articles}\tokenH{ in}\tokenH{ the}\tokenH{ three}\tokenH{ days}\tokenH{ is}\tokenH{ \textbackslash{}(}\tokenH{\textbackslash{}}\tokenH{boxed}\tokenH{\{}\tokenH{1}\tokenH{0}\tokenH{4}\tokenH{\}}\tokenH{\textbackslash{})}\tokenH{.}
\end{tcolorbox}

\begin{tcolorbox}[colback=white, colframe=gray!80, title=LLaDA Prediction with PVF (36 steps), sharp corners=south]
\tokenH{To}\tokenH{ determine}\tokenH{ the}\tokenH{ total}\tokenH{ number}\tokenH{ of}\tokenH{ hours}\tokenH{ Meredith}\tokenH{ spent}\tokenH{ writing}\tokenH{ articles}\tokenF{ over}\tokenH{ the}\tokenH{ three}\tokenH{ days}\tokenH{,}\tokenH{ we}\tokenH{ need}\tokenH{ to}\tokenH{ follow}\tokenH{ these}\tokenH{ steps}\tokenP{:}
\\
\\
\tokenH{**}\tokenP{Step}\tokenH{ 1}\tokenH{:}\tokenH{ Calculate}\tokenH{ the}\tokenH{ number}\tokenH{ of}\tokenH{ articles}\tokenH{ written}\tokenF{ on}\tokenF{ Tuesday}\tokenH{.**}
\\
\\
\tokenH{-}\tokenH{ On}\tokenH{ Monday}\tokenF{,}\tokenH{ Meredith}\tokenH{ wrote}\tokenH{ 5}\tokenH{ articles}\tokenH{.}
\\
\tokenH{-}\tokenH{ On}\tokenH{ Tuesday}\tokenF{,}\tokenH{ she}\tokenH{ wrote}\tokenH{ \textbackslash{}(}\tokenH{\textbackslash{}frac}\tokenP{\{}\tokenH{2}\tokenH{\}}\tokenH{\{}\tokenH{5}\tokenH{\}}\tokenH{\textbackslash{})}\tokenH{ times}\tokenH{ more}\tokenH{ articles}\tokenH{ than}\tokenH{ on}\tokenH{ Monday}\tokenF{.}
\\
\tokenF{-}\tokenH{ To}\tokenF{ find}\tokenH{ the}\tokenH{ number}\tokenH{ of}\tokenH{ articles}\tokenH{ written}\tokenH{ on}\tokenH{ Tuesday}\tokenP{,}\tokenH{ we}\tokenH{ calculate}\tokenP{ \textbackslash{}(}\tokenH{\textbackslash{}frac}\tokenH{\{}\tokenH{2}\tokenH{\}}\tokenH{\{}\tokenH{5}\tokenP{\}}\tokenF{\textbackslash{})}\tokenH{ of}\tokenF{ 5}\tokenH{:}
\\
\tokenH{\textbackslash{}[}
\\
\tokenH{\textbackslash{}frac}\tokenP{\{}\tokenH{2}\tokenH{\}}\tokenH{\{}\tokenH{5}\tokenH{\}}\tokenH{ \textbackslash{}times}\tokenH{ 5}\tokenP{ =}\tokenH{ 2}
\\
\tokenH{\textbackslash{}]}
\\
\tokenP{-}\tokenH{ Therefore}\tokenP{,}\tokenH{ the}\tokenP{ number}\tokenH{ of}\tokenH{ articles}\tokenH{ written}\tokenH{ on}\tokenH{ Tuesday}\tokenH{ is}\tokenH{:}
\\
\tokenH{\textbackslash{}[}
\\
\tokenH{5}\tokenH{ +}\tokenH{ 2}\tokenH{ =}\tokenH{ 7}
\\
\tokenH{\textbackslash{}]}
\\
\\
\tokenH{**Step}\tokenH{ 2}\tokenH{:}\tokenH{ Calculate}\tokenH{ the}\tokenH{ number}\tokenH{ of}\tokenH{ articles}\tokenH{ written}\tokenH{ on}\tokenH{ Wednesday}\tokenH{.**}
\\
\\
\tokenH{-}\tokenH{ On}\tokenH{ Wednesday}\tokenF{,}\tokenH{ she}\tokenH{ wrote}\tokenH{ twice}\tokenH{ the}\tokenH{ number}\tokenH{ of}\tokenH{ articles}\tokenH{ she}\tokenH{ wrote}\tokenH{ on}\tokenH{ Tuesday}\tokenH{.}
\\
\tokenH{-}\tokenH{ On}\tokenH{ Tuesday}\tokenP{,}\tokenH{ she}\tokenH{ wrote}\tokenH{ 7}\tokenH{ articles}\tokenH{.}
\\
\tokenH{-}\tokenH{ Therefore}\tokenH{,}\tokenH{ the}\tokenH{ number}\tokenH{ of}\tokenH{ articles}\tokenH{ written}\tokenH{ on}\tokenH{ Wednesday}\tokenH{ is}\tokenH{:}
\\
\tokenH{\textbackslash{}[}
\\
\tokenH{2}\tokenH{ \textbackslash{}times}\tokenH{ 7}\tokenP{ =}\tokenH{ 14}
\\
\tokenH{\textbackslash{}]}
\\
\\
\tokenH{**Step}\tokenH{ 3}\tokenH{:}\tokenH{ Calculate}\tokenH{ the}\tokenH{ total}\tokenH{ number}\tokenH{ of}\tokenH{ articles}\tokenH{ written}\tokenH{ over}\tokenH{ the}\tokenH{ three}\tokenH{ days}\tokenH{.**}
\\
\\
\tokenP{-}\tokenH{ On}\tokenH{ Monday}\tokenP{,}\tokenH{ she}\tokenF{ wrote}\tokenH{ 5}\tokenH{ articles}\tokenH{.}
\\
\tokenH{-}\tokenH{ On}\tokenH{ Tuesday}\tokenH{,}\tokenH{ she}\tokenH{ wrote}\tokenH{ 7}\tokenH{ articles}\tokenH{.}
\\
\tokenH{-}\tokenH{ On}\tokenH{ Wednesday}\tokenH{,}\tokenH{ she}\tokenH{ wrote}\tokenH{ 14}\tokenH{ articles}\tokenH{.}
\\
\tokenH{-}\tokenH{ The}\tokenH{ total}\tokenH{ number}\tokenP{ of}\tokenH{ articles}\tokenH{ written}\tokenH{ over}\tokenH{ the}\tokenH{ three}\tokenH{ days}\tokenH{ is}\tokenH{:}
\\
\tokenH{\textbackslash{}[}
\\
\tokenH{5}\tokenH{ +}\tokenH{ 7}\tokenP{ +}\tokenH{ 14}\tokenH{ =}\tokenH{ 26}
\\
\tokenH{\textbackslash{}]}
\\
\\
\tokenH{**Step}\tokenH{ 4}\tokenH{:}\tokenH{ Calculate}\tokenH{ the}\tokenH{ total}\tokenH{ number}\tokenH{ of}\tokenH{ hours}\tokenH{ spent}\tokenH{ writing}\tokenH{ articles}\tokenH{.**}
\\
\\
\tokenH{-}\tokenH{ Each}\tokenH{ article}\tokenH{ takes}\tokenH{ an}\tokenH{ average}\tokenP{ of}\tokenH{ 4}\tokenH{ hours}\tokenH{ to}\tokenH{ research}\tokenH{ and}\tokenH{ write}\tokenH{.}
\\
\tokenH{-}\tokenH{ Therefore}\tokenH{,}\tokenH{ the}\tokenH{ total}\tokenH{ number}\tokenH{ of}\tokenH{ hours}\tokenH{ spent}\tokenH{ writing}\tokenH{ articles}\tokenH{ is}\tokenH{:}
\\
\tokenH{\textbackslash{}[}
\\
\tokenH{26}\tokenP{ \textbackslash{}times}\tokenH{ 4}\tokenH{ =}\tokenH{ 1}\tokenF{04}
\\
\tokenH{\textbackslash{}]}
\\
\\
\tokenH{Thus}\tokenH{,}\tokenH{ the}\tokenH{ total}\tokenH{ number}\tokenH{ of}\tokenH{ hours}\tokenH{ Meredith}\tokenH{ spent}\tokenH{ writing}\tokenH{ articles}\tokenF{ in}\tokenH{ the}\tokenH{ three}\tokenH{ days}\tokenH{ is}\tokenP{ \textbackslash{}(}\tokenH{\textbackslash{}boxed}\tokenH{\{}\tokenH{104}\tokenH{\}}\tokenH{\textbackslash{})}\tokenH{.}
\end{tcolorbox}

\newpage
\begin{algorithm}[t]
\footnotesize
\caption{Plan-Verify-Fill (PVF) Decoding}
\label{alg:pvf}
\begin{algorithmic}[1]
\State \textbf{Input:} model $P_\theta$, length $L$, thresholds $\tau_{\mathrm{high}},\tau_{\mathrm{plan}}$, width $K$
\State \textbf{Initialize:} $\mathbf{y}_0\leftarrow[\texttt{MASK}]^L$, $t\leftarrow0$

\While{$\mathbf{y}_t$ contains \texttt{[MASK]}}
    \State Compute top predictions and confidences on the current active block.
    \State Construct the high-confidence base branch $\mathbf{z}^{\mathrm{base}}_t$.
    \State Identify planning candidates $\mathcal{P}^{\mathrm{plan}}_t$.
    \State Set $\mathcal{J}^{\mathrm{plan}}_t\leftarrow\emptyset$.

    \If{$\mathcal{P}^{\mathrm{plan}}_t\neq\emptyset$}
        \State \textbf{Planning Route.}
        \State Construct up to $K$ planning branches $\{\mathbf{z}^{\mathrm{plan},j}_t\}_{j=1}^{K_t}$.
        \State Run one batched forward pass on $\{\mathbf{z}^{\mathrm{base}}_t,\mathbf{z}^{\mathrm{plan},1}_t,\ldots,\mathbf{z}^{\mathrm{plan},K_t}_t\}$.
        \State Apply impact-set verification.
        \State Let $\mathcal{J}^{\mathrm{plan}}_t$ be the set of verified planning branches.
        \If{$\mathcal{J}^{\mathrm{plan}}_t\neq\emptyset$}
            \State Select $j^*$ among verified planning branches.
            \State Commit $\mathbf{y}_{t+1}\leftarrow\mathbf{z}^{\mathrm{plan},j^*}_t$.
        \Else
            \State Commit $\mathbf{z}_t^{\mathrm{base}}$.
        \EndIf
    
    \Else
        \State \textbf{AR Fallback Route.}
        \State Construct nested fallback branches $\{\mathbf{z}^{\mathrm{AR},k}_t\}_{k=1}^{K_t^{\mathrm{AR}}}$.
        \State Verify fallback branches by the local consistency.
        \If{no fallback branch passes verification}
            \State Commit $\mathbf{y}_{t+1}\leftarrow\mathbf{z}^{\mathrm{base}}_t$.
        \Else
            \State Let $k^*$ be the largest verified fallback length.
            \State Commit $\mathbf{y}_{t+1}\leftarrow\mathbf{z}^{\mathrm{AR},k^*}_t$.
        \EndIf
    \EndIf

    \State $t\leftarrow t+1$
\EndWhile

\State \textbf{Return} $\mathbf{y}_t$
\end{algorithmic}
\vspace{-0.1cm}
\end{algorithm}
\end{document}